\documentclass{siamart220329}
\usepackage[utf8]{inputenc}
\usepackage{tikz}
\usepackage{graphicx,float}
\usepackage{bigints}
\usepackage{subcaption}
\usepackage{hyperref}
\usepackage{multirow}
\usepackage{textcomp}
\usepackage{array}
\usepackage{ragged2e}
\usepackage{arydshln}
\usepackage{lineno}
\usepackage{bm}
\usepackage{enumerate}
\usepackage{cite}

\usepackage{placeins}
\usepackage{flafter} 

\usepackage{makecell}

\makeatletter
\renewcommand*\env@matrix[1][\arraystretch]{%
  \edef\arraystretch{#1}%
  \hskip -\arraycolsep
  \let\@ifnextchar\new@ifnextchar
  \array{*\c@MaxMatrixCols c}}
\makeatother

\usepackage{tikz}
\usetikzlibrary{shapes.geometric, arrows, calc, fit}
\usetikzlibrary{backgrounds}
\tikzstyle{arrow} = [thick,->,>=stealth] 

\usepackage[ruled,lined, linesnumbered,longend, algo2e]{algorithm2e}
\SetKwInput{KwInput}{Input}
\SetKwInput{KwInitialize}{Initialize}

\let\oldnl\nl
\newcommand{\nonl}{\renewcommand{\nl}{\let\nl\oldnl}}%

\makeatletter
\newcounter{Function}
\newenvironment{Function}[1][htb]
  {
   \let\c@algocf\c@Function
   \begin{algorithm2e}[#1]%
  }{\end{algorithm2e}}
\makeatother

\usepackage{amsmath,amsfonts,amssymb,mathtools}

\newsiamremark{notation}{Notation}
\newsiamremark{remark}{Remark}

\newcommand{\ip}[2]{\left \langle #1,\ #2\right \rangle}

\headers{weak-SINDy compression}{Russo, Laiu, Archibald}

\title{Streaming Compression of Scientific Data via weak-SINDy\thanks{This manuscript has been authored by UT-Battelle, LLC, under contract DE-AC05-00OR22725 with the US Department of Energy (DOE). The US government retains and the publisher, by accepting the work for publication, acknowledges that the US government retains a non-exclusive, paid-up, irrevocable, world-wide license to publish or reproduce the submitted manuscript version of this work, or allow others to do so, for US government purposes. DOE will provide public access to these results of federally sponsored research in accordance with the DOE Public Access Plan (http://energy.gov/downloads/doe-public-access-plan).}}

\author{Benjamin P. Russo\thanks{Riverside Research, New York, NY 10038, USA. (\email{brusso@riversideresearch.org})}
\and M. Paul Laiu\thanks{Mathematics in Computation Section, Computer Science and Mathematics Division, Oak Ridge National Laboratory, Oak Ridge, TN 37831, USA.
  (\email{laiump@ornl.gov}, \email{archibaldrk@ornl.gov}.)} \and Richard Archibald\footnotemark[3]}

\newcommand{\changed}[1]{{\color{blue}\ \\ This has been changed.}}

\begin{document}

\maketitle
\begin{abstract}
In this paper a streaming weak-SINDy algorithm is developed specifically for compressing streaming scientific data. The production of scientific data, either via simulation or experiments, is undergoing a stage of exponential growth, which makes data compression important and often necessary for storing and utilizing large scientific data sets. As opposed to classical ``offline" compression algorithms that perform compression on a readily available data set, streaming compression algorithms compress data ``online" while the data generated from simulation or experiments is still flowing through the system. This feature makes streaming compression algorithms well-suited for scientific data compression, where storing the full data set offline is often infeasible.
This work proposes a new streaming compression algorithm, streaming weak-SINDy, which takes advantage of the underlying data characteristics during compression. The streaming weak-SINDy algorithm constructs feature matrices and target vectors in the online stage via a streaming integration method in a memory efficient manner. The feature matrices and target vectors are then used in the offline stage to build a model through a regression process that aims to recover equations that govern the evolution of the data.
For compressing high-dimensional streaming data, we adopt a streaming proper orthogonal decomposition (POD) process to reduce the data dimension and then use the streaming weak-SINDy algorithm to compress the temporal data of the POD expansion. We propose modifications to the streaming weak-SINDy algorithm to accommodate the dynamically updated POD basis. By combining the built model from the streaming weak-SINDy algorithm and a small amount of data samples, the full data flow could be reconstructed accurately at a low memory cost, as shown in the numerical tests.
\end{abstract}



\section{Introduction}
With the increase in computing power, there has been a substantial increase in data generated by experiments and simulations. For example, climate models often produce peta-bytes of data \cite{climate}, making storage and analysis of the data difficult. Moreover, at high performance computing and experimental facilities, data is generated without the ability to store and revisit past entries, which is spurring online techniques. 
This drives the development of online dictionary learning techniques \cite{archibald2022dictionary, online_dictionary}. 
However, scientific data are often governed by underlying physical systems, which are usually described using ordinary differential equations (ODEs) or partial differential equations (PDEs). 
These governing equations give a compact representation of the scientific data, i.e., the data can be fully recovered by solving these equations when system configurations, such as initial and boundary conditions, are known.
In the scenarios that the governing equations are either unknown or expensive to solve, which are often the case for underlying systems for massive, high-dimensional scientific data, reconstructing data by solving the ODEs or PDEs is infeasible.
This motivates the development of techniques that take advantage of some inexpensive approximations, e.g., surrogate models, for the underlying systems to improve the efficiency of the compression and recovery of streaming scientific data. 
Recent advances in non-linear system identification \cite{brunton2016discovering,messenger2021weak,messenger2021weakpde,SCC.Rosenfeld.Kamalapurkar.ea2019a} offer simple yet effective data-driven techniques to recover governing equations. In these SINDy (Sparse Identification of Non-linear Dynamics) type techniques, recovery of the governing equations is framed as solving parameter identification problems via linear regression. Moreover, it was shown in \cite{russo2022convergence} that weak formulations of SINDy can generate accurate surrogate models. 
In this article, we develop a streaming weak-SINDy algorithm for streaming data compression, which can be combined with a streaming proper orthogonal decomposition (POD) method to target applications in high-dimensional scientific data compression.

\paragraph{Problem Statement} We assume that the incoming data is a stream of snapshots $\{\bm{u}(t_n)\}^N_{n=1}(=:\{\bm{u}_n\}^N_{n=1})$ of a solution $\bm{u}:[0,T]\rightarrow \mathbb{R}^S$ to an ODE
\begin{equation}
\label{eq:gen_ODE_system}
\frac{d\bm{u}}{dt} = f(\bm{u}(t)).
\end{equation}
The goal is to perform data compression by learning a surrogate model for $f(\bm{u}(t))$ from streaming data in real-time and to use the learned model in the offline stage for data decompression. In this, although some storage is allocated for the online process, we assume that in general the entire data set $\{\bm{u}_n\}_{n=1}^N$ is too large to store for offline compression. 

Overall, this compression technique takes advantage of the form of scientific streaming data and is orthogonal to standard approaches such as compressed sensing \cite{streaming_compressed_sensing} or traditional video compression \cite{video_compression}, which are agnostic to the underlying structure of the data. In particular, standard approaches do not take advantage of the overall temporal patterns of the data. In contrast, the proposed streaming weak-SINDy algorithm builds a relatively inexpensive model that captures the temporal evolution of the data and leverages it in the compression/decompression process. Hence, an advantage is gained in situations where a simulation is expensive or where the dynamics are entirely unknown. 

There exist some examples in the literature of the online generation of surrogate models such as an online version of dynamic mode decomposition (DMD)\cite{streaming_DMD, zhang2019online} and online weak-SINDy (a Galerkin approach to SINDy) \cite{messenger2022online}. In these approaches, the model is updated in real time as more data becomes available. Despite being computationally efficient, streaming variants of DMD proposed in \cite{streaming_DMD, zhang2019online} inherit the disadvantages of DMD \cite{wu2021challenges, baddoo2022kernel}. Most notably, they fundamentally result in linear models which poorly predict dynamics with strong nonlinearity. Additionally, DMD models are typically sensitive to noise. 
On the other hand, the online weak-SINDy approach in \cite{messenger2022online} updates the resulting nonlinear surrogate models as more data becomes available by solving linear systems that are constructed via applying a convolution-based technique on a moving window of data. 

In this work, we aim to compress the streaming data using a surrogate model that is built after data collection rather than to construct online surrogate models. Therefore, we take a different approach than the one in \cite{messenger2022online}. Specifically, we update the components of the linear system via a streaming integration process and save the least squares solve for an offline process. Since the size of the linear system is independent of the number of incoming streaming data, this updating process has a much smaller memory footprint than the standard SINDy methods.
For high dimensional systems, we incorporate a streaming POD technique (see, e.g., \cite{streaming_POD} and \cite{streaming_SPOD}) to reduce the data dimension by constructing POD expansion on a dynamically updated basis. The steaming POD technique we adapted from \cite{streaming_POD} computes an initial POD basis from a small subset of data and adds new components to the basis when the incoming data cannot be well-approximated by the original basis. We propose modifications to the streaming weak-SINDy algorithm to accommodate the varying size of a POD basis, in which each basis update results in an augmentation in the linear system. The augmented linear system is then solved in the offline stage to build models that serve as a compressed form of the updated temporal POD modes.
By combining streaming weak-SINDy with the streaming POD algorithm, we offer an algorithm for compressing high dimensional streaming scientific data with a small online memory requirement. 

In Section \ref{sec:SINDy_review}, we cover the necessary pre-requisites for this paper by giving a brief overview of SINDy, its integral extension weak-SINDy, and POD. In Section \ref{sec:Streaming_weak_SINDy_for_compression}, we outline the proposed streaming weak-SINDy algorithm. Section \ref{sec:high-dimensional_data_compression} introduces the combination of the streaming weak-SINDy algorithm and a streaming POD approach for compressing high-dimensional streaming scientific data. Section \ref{sec:Experiments} displays proof-of-concept numerical examples for compressing data from a low-dimensional system, the Lorenz system, and a high-dimensional fluid-flow data example. 
For clarity, we set notation for a list of parameters consistent throughout the article. 

\begin{notation}\ \\
\begin{center}
\renewcommand{\arraystretch}{1} 
\begin{tabular}{|l|l|l|}
\hline
 $S\colon$ state dimension 
& $N\colon$ number of time steps\\
\hline
 $K\colon$ number of test functions
& $J\colon$ number of basis functions\\
\hline
 $L\colon$ number of POD modes
&$M\colon$ number of added POD modes\\
\hline
$P\colon$ degree of the Newton-Cotes formula & \ \\
\hline
\end{tabular}
\end{center}
\end{notation}\ \\
Some additional notational conventions we will follow throughout the paper include:
\begin{enumerate}[--]
\item bold lower case $\bm{v}$ for vector valued objects, 
\item bold upper case $\bm{A}$ for matrix valued objects,
\item if a parameter ``$Q$" represents a number of objects, we will index by the lower case letter ``$q$",
\item snapshots $\bm{u}(t_n)$ may be abbreviated by $\bm{u}_n$,
\item as all objects are finite sets of vectors and/or matrices, the sum of the $L_0$ norms will serve as a measure of the size of these sets and will be denoted $\mu(\cdot )$.
\end{enumerate}

\section{Background material}
\label{sec:SINDy_review}

In this section, we give a brief overview for the two main components in the proposed streaming compression algorithm -- the weak-SINDy method and the POD technique.

\subsection{The weak-SINDy method}
\label{subsec:weak-SINDy}

The original SINDy method \cite{brunton2016discovering} aims to identify an underlying dynamical system
$\dot{u}= f(u(t))$ from measured or simulated solution data $u$ in a time interval $[0,T]$. Here $\dot{u}$ denotes the time derivative of $u$, which will be used throughout this paper. In SINDy, functions evaluated on the snapshots of the trajectory are used as feature vectors and derivative snapshots are used as a target variable. Features are then fitted to the target via a sparsity promoting technique such as LASSO \cite{tibshirani1996regression}. In this section, we consider the case when $u$ is a scalar-valued function for simplicity. The extension of SINDy (and weak-SINDy) to vector-valued functions is straightforward by a component-wise application (see, e.g., \cite{brunton2016discovering}).

\paragraph{SINDy}
Given $\{u(t_k)\}_{k=1}^K$ at time steps $\{t_k\}_{k=1}^K$, the original SINDy method \cite{brunton2016discovering} seeks a sparse set of coefficients $\{c_j\}_{j=1}^J$ that satisfy
\begin{equation}\textstyle\dot{u}(t_k) = \sum_{j=1}^J c_j \varphi_j(u(t_k)),\quad k=1,\dots,K,\end{equation}
which is then formulated as a linear system
\begin{equation}\label{eq:SINDy}
    \bm{b} = \bm{G}\bm{c} \quad\text{with entries}\quad
 \bm{b}_k = \dot{u}(t_k),\quad
 \bm{G}_{k,j} = \varphi_j(u(t_k)),\quad\text{and}\quad
 \bm{c}_j = c_j.
\end{equation}
Despite being shown to be effective for identifying equations in several applications, the SINDy method requires the time-derivative data in the target vector $\bm{b}$ and thus suffers from noise sensitivity in time-derivative measurements or estimations.
 
To address this, integral formulations of SINDy have been created. Several techniques to the author's knowledge are Messenger and Bortz's weak formulation \cite{messenger2021weak, messenger2021weakpde}, Rosenfeld et~al.'s operator theoretic technique \cite{SCC.Rosenfeld.Kamalapurkar.ea2019a, rosenfeld2019occupation}, the integral-term-based model selection technique proposed by Schaeffer and McCalla \cite{PhysRevE.96.023302}, and the work of Gurevich et~al. in \cite{Gurevich}. Below, we introduce weak-SINDy as that is the main focus of this article.

\paragraph{weak-SINDy}
The weak-SINDy method proposed in \cite{messenger2021weak, messenger2021weakpde} introduces a class of \emph{test} functions $\{\psi_k\}_{k=1}^K$ in addition to the basis functions $\{\varphi_j\}_{j=1}^J$ in the SINDy method. Throughout this paper, we refer to $\{\varphi_j\}_{j=1}^J$ as a \emph{projection basis} and $\{\psi_k\}_{k=1}^K$ as a \emph{test function basis}. Given $u(t)$, $\forall t\in[0,T]$, the weak-SINDy method finds the weights $\{c_j\}_{j=1}^J$ by solving
\begin{equation}\label{eq:wSINDy}
\textstyle
    \ip{\dot{u}}{\psi_k} = \ip{\sum_{j=1}^J c_j\varphi_j(u)}{\psi_k},\quad k=1,\dots,K,
\end{equation}  
where $\ip{\cdot}{\cdot}$ denotes the $L^2$ inner product on $[0,T]$.
In matrix form, Eq.~\eqref{eq:wSINDy} is written as 
\begin{equation}
\label{eq:weak-SINDy_form}
\bm{b} = \bm{G} \bm{c}\quad\text{with entries}\quad
 \bm{b}_k = \ip{\dot{u}}{\psi_k}\quad\text{and}\quad
 \bm{G}_{k,j} = \ip{\varphi_j(u)}{\psi_k},
\end{equation}
where the entries of $\bm{b}$ can be computed using integration by parts, i.e.,
\begin{equation}\label{eq:integration_by_parts}
\textstyle
\ip{\dot{u}}{\psi_k} = -\langle{u},{\dot{\psi_k}}\rangle,
\end{equation}
under the assumption that $\{\psi_k\}_{k=1}^K$ have compact support in $[0,T]$.
Therefore, the weak-SINDy method avoids direct computations of $\dot{u}$ by invoking $\dot{\psi_k}$, the derivatives of the test function, which is often known a priori. Moreover, model selection based on integral terms performs well in the presence of noise \cite{PhysRevE.96.023302}. In lieu of using compactly supported test functions, one can use any class of test functions by incorporating boundary data. In the one dimensional case, 
\begin{equation}\label{eq:integration_by_parts-noncompact_supp}
\textstyle
\ip{\dot{u}}{\psi_k} = -\langle{u},{\dot{\psi_k}}\rangle + \left.u\cdot \psi_k\right|^T_{0}.
\end{equation}
Although the weak-SINDy method was developed to address the issue on the time-derivative measurement or estimation noises, we show in Section~\ref{sec:Streaming_weak_SINDy_for_compression} that the integral form of weak-SINDy allows for applications to streaming data.

For the rest of this paper, we shall refer to $\bm{b}$ as the \textit{target vector}, $\bm{G}$ as the \textit{feature matrix}, and $\bm{c}$ as the \textit{coefficient vector}. In a multi-variable setting, we still denote the matrices whose columns are the target vectors and coefficient vectors as $\bm{b}$ and $\bm{c}$, respectively. When we do need to explicitly denote a target vector or coefficient vector corresponding to a particular component, we will again use an index. In a multi-variable system with $S$ components, the target vectors and coefficient vectors are indexed as $\bm{b}_{s}$ and $\bm{c}_s$, respectively, with $s=1,\dots,S$. An example of this can be seen in Section \ref{sec:high-dimensional_data_compression}.

\subsection{Proper orthogonal decomposition (POD)}
\label{subsec:POD}

The goal of POD is to efficiently represent a function $u(\bm{x},t)$ in a finite number of modes so that
\[u(\bm{x},t) \approx \sum_{\ell=1}^L \nu_\ell(t)\upsilon_\ell(\bm{x}).\]
Here, we assume that $\bm{x}\in \Omega\subset\joinrel\subset \mathbb{R}^d$, $\Omega$ compact, and $t\in [0,T]$. In this section, we call $\{\upsilon_\ell\}_{\ell=1}^L$ the \textit{spatial modes} and $\{\nu_\ell\}_{\ell=1}^L$ the \textit{temporal modes}. 
Consider an operator kernel $R(\bm{x},\bm{y})$ defined as 
\begin{equation}\label{eq:R-formula}
R(\bm{x},\bm{y}):=\frac{1}{T}\int_0^T u(\bm{x},t)u(\bm{y},t)\,dt\:,
\end{equation}
which can be interpreted as the time-average of $u(\bm{x},t) u(\bm{y},t)$. If $R(\bm{x},\bm{y})$ is continuous and thus square integrable over the compact domain $\Omega\times \Omega$, then $R$ is the kernel of a Hilbert-Schmidt integral operator. Then, the operator $\mathcal{R}$ defined as
\begin{equation}\label{eq:R-operator_formula}
[\mathcal{R}w](\bm{x}):=\int_{\Omega} R(\bm{x},\bm{y}) w(\bm{y}) \,d\bm{y} 
\end{equation}
is self-adjoint since $R(\bm{x},\bm{y}) = R(\bm{y},\bm{x})$. Therefore, the spectral theorem guarantees that there exists an orthonormal basis $\{\upsilon_\ell(\bm{x})\}_{\ell=1}^\infty$ of eigenfunctions of $\mathcal{R}$, which are considered as the spatial modes of $u(\bm{x},t)$. 
By orthogonality of $\{\upsilon_\ell(\bm{x})\}_{\ell=1}^\infty$, the temporal modes can be defined as
\begin{equation}\label{eq:temporal-modes}
\nu_\ell(t) = \int_\Omega u(\bm{x},t) \upsilon_\ell(\bm{x})\, dx = \langle u(\bm{x},t),\upsilon_\ell(\bm{x})\rangle_\Omega.
\end{equation}
where the inner product is taken over $\Omega$. 

We note that a truncated POD expansion minimizes the average of the spatial $L^2$ error over time. Let $u_\text{pod}(\bm{x},t) = \sum_{\ell = 1}^L \nu_\ell(t) \upsilon_\ell(\bm{x})$ be a truncation of the expansion of $u(\bm{x},t)$ in terms of the POD basis. Now the time average of the $L^2$ spatial norm error can be bounded by
\begin{equation}
\begin{alignedat}{2}
\label{eq:spectral_error}
\textstyle
\frac{1}{T}\int_{0}^T \|u(\bm{x}, t) - u_\text{pod}(\bm{x},t)\|_\Omega\, dt
&\textstyle
 = \frac{1}{T}\int_{0}^T \left\|\sum^\infty_{\ell = L+1} \nu_\ell(t) \upsilon_\ell(\bm{x})\right\|_\Omega\, dt\\
&\textstyle
 = \frac{1}{T}\int_{0}^T \sqrt{\sum^\infty_{\ell = L+1}|\nu_\ell(t)|^2}\, dt\\
&\textstyle
 \leq \sum^\infty_{\ell = L+1} \frac{1}{T}\int_{0}^T|\nu_\ell(t)|\, dt =  \sum^\infty_{\ell = L+1} \lambda_\ell,
\end{alignedat}
\end{equation}
where $\lambda_\ell$ denotes the $\ell$-th eigenvalue of the integral operator \eqref{eq:R-operator_formula}. Hence,the decay rate of the eigenvalues correlates to the efficiency of the POD approximation $u_\text{pod}(\bm{x},t)$.
Finally, by applying the following theorem from \cite[Theorem~2]{eigenvaluedecay}, it can be shown that the overall regularity of the data $u(\bm{x},t)$ has a significant effect on the efficiency of POD. 
\begin{theorem}[Chang and Ha]
\label{thm:ChangandHa}
If $K(x,y)$ is a positive definite Hermitian kernel such that the partial derivatives $\frac{\partial^p K(x,y)}{\partial y^p}K(x,y)$ exists and is continuous on $[0,1]^2$.
\begin{equation}
\label{eq:eigen_decay_rate}
\lambda_\ell = o(1/\ell^{p+1}) \quad\text{ as }\quad \ell \rightarrow \infty.
\end{equation}
Here $\lambda_\ell$ denotes the eigenvalues of $[\mathcal{K}w]({x}):=\int_{\Omega} K({x},{y}) w({y}) \,d{y}$ arranged in decreasing order. 
\end{theorem}
Specifically, the kernel $R$ in Eq.~\eqref{eq:R-formula} is indeed a positive definite symmetric kernel. By applying Theorem~\ref{thm:ChangandHa} to $R$, it follows that an increasing  regularity of $u$ (and thus $R$) corresponds to a faster decay rate for the eigenvalues of $\mathcal{R}$, resulting a lower error as shown in Equation~\eqref{eq:spectral_error}.

\paragraph{Snapshot method for generating POD modes}
\label{par:Snapshot_method} In this paper, the spatial POD modes are calculated via the method of snapshots (see, e.g., \cite{holmes2012turbulence}). We include a brief overview for completeness. Given a discretization $\hat{\bm{x}} = [x_1, \ldots, x_S]$ of the domain $\Omega$ we let $\bm{u}(\hat{\bm{x}}, t_n) = [u(x_1, t_n), \ldots u(x_S,t_n)]^\top\in \mathbb{R}^S$ be a snapshot of $u(\bm{x},t)$ at $t_n$. Suppose the data ${D}$ is a collection of these snapshots from a discretization of the domain $[0,T]$, i.e., $D = \{\bm{u}(\hat{\bm{x}}, t_1), \ldots \bm{u}(\hat{\bm{x}},t_N)\}\subset \mathbb{R}^{S}$. Forming the data matrix $\bm{D}=[\bm{u}(\hat{\bm{x}}, t_1), \ldots \bm{u}(\hat{\bm{x}},t_N)]\in \mathbb{R}^{S\times N}$, we compute the singular value decomposition $[\bm{U}, \bm{\Sigma}, \bm{V}]= \operatorname{svd}(\bm{D})$. Here the columns of $\bm{U}$ contain the discretized spatial modes and the rows of $\bm{V}$ contain the (scaled) discretized temporal modes. We also assume that the columns are organized according to the magnitude of the singular values in descending order. 

\section{Streaming weak-SINDy for compression}
\label{sec:Streaming_weak_SINDy_for_compression}
Our goal is to compress a data stream of ODE data $\{\bm{u}_n\}^N_{n=1}$, with each snapshot $\bm{u}_n\in \mathbb{R}^S$. To do so, we seek to develop a method of creating surrogate models with low online storage requirements in comparison to the storage requirements of the entire data set. For a given data set $\{\bm{u}_n\}$, to construct a weak-SINDy surrogate model, we use a set of test function $\{\psi_k\}^K_{k=1}$ and a set of projection functions $\{\varphi_j\}^J_{j=1}$. Then the feature matrix $\bm{G}$ and target vectors $\{\bm{b}_s\}^S_{s=1}$ (one target vector for each component of $\bm{u}_n\in \mathbb{R}^S$) are constructed with the entries given by the integrals in Eqs.~\eqref{eq:wSINDy} and \eqref{eq:weak-SINDy_form}. Note that the sizes of the target vectors and feature matrix are determined by the number of test functions and the number of basis functions and does \emph{not} depend on the number of time steps, which makes weak-SINDy particularly suitable for applications to streaming data, where the number of time steps are often large and not known a priori. In general, the size of the feature matrix and target vectors is much smaller than the size of the data set. After the streaming integration is done, we can perform a regression to further compress and then discard the feature matrix and target vectors. 

\begin{figure}
\begin{center}
\scalebox{.87}{%
\includegraphics[]{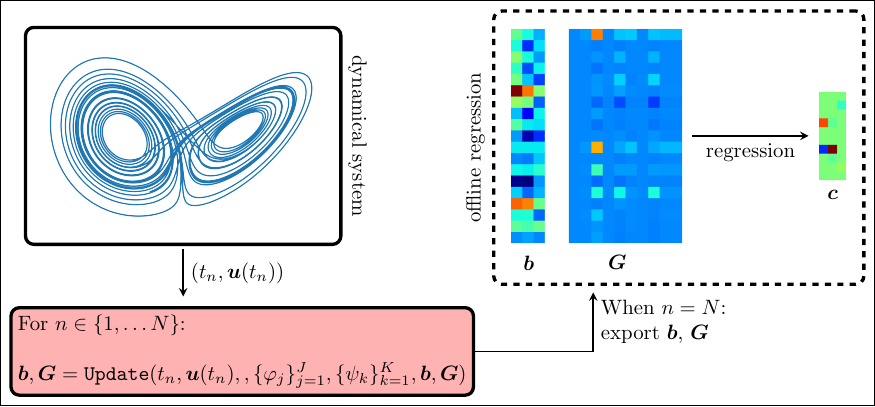}
}
\end{center}
\caption{A diagram of the streaming weak-SINDy algorithm}
\label{fig:streaming_weak_SINDy_diagram}
\end{figure}

In Section \ref{subsec:streaming_integration}, we outline the details of a streaming integration procedure which is used to update the entries of the feature matrix and target vectors. We then give an overview and pseudo code (Algorithm~\ref{alg:streaming_weak_SINDy}) for the overall process in Section \ref{subsec:streaming_weak-SINDy_alg}. A graphical summary of Algorithm \ref{alg:streaming_weak_SINDy} is presented in Figure \ref{fig:streaming_weak_SINDy_diagram}. 

\subsection{Streaming integration}
\label{subsec:streaming_integration}

Under the assumption of equally spaced snapshots, we first outline a simple streaming integration method based on any composite rule from a Newton-Cote's formula.  

\paragraph{Streaming integration with uniform time intervals:}
\label{par:streaming_int_equal_spacing}
For a collection of $N$ snapshots $\{g(t_1), \ldots ,g(t_N)\}$, where the distance between successive points is a fixed value $\Delta t = |t_{i+1} - t_{i}|$, a composite integration rule using a $P$-th degree Newton-Cotes formula has the form
\begin{equation}
\label{eq:Newton_Cotes_composite_rule}
\int_0^T g(t)\, dt \approx \sum_i \alpha_P \cdot \Delta t \cdot (w_1 g_{i,1} + w_2 g_{i,2}+ \ldots+ w_P g_{i,P}).
\end{equation}
Here, $\{g_{i,1}, g_{i, 2}, \ldots, g_{i,P}\}$ represents the $i$-th set of $P$-many snapshots of the function and $\{\alpha_P, w_1, \ldots w_P\}$ represents the set of weights for the $P$-th degree Newton-Cotes formula. It is important to note that the integral value can be updated for every set of $P$ snapshots recorded.  Let $I(i)$ represent the value of the integral after the $i$-th set of $P$ many snapshots  $\{g_{i,1}, g_{i, 2}, \ldots, g_{i,P}\}$. This motivates the update scheme, given the $i+1$-th set of snapshots $\{g_{i+1,1}, \ldots, g_{i+1,P'}\}$, of size $P'$, then
\begin{equation}
\label{eq:int_scheme1}
I(i+1) = \left\{\begin{array}{ll}
I(i)  +\alpha_{P} \cdot \Delta t \cdot (w_1 g_{i+1,1} + \ldots+ w_{P} g_{i+1,P'}) 
& \text{ if } P' = P\\ 
I(i)  +\alpha_{P'} \cdot \Delta t \cdot (w_1 g_{i+1,1} + \ldots+ w_{P'} g_{i+1,P'}) & \text{ if } P'<P\\ 
\end{array}
\right..
\end{equation}
In the above, if not enough snapshots are collected then a lower ($P{'}$-th) degree Newton-Cotes formula is substituted. Moreover, the first $P-1$ snapshots may be discarded after each iteration and the $P$-th snapshot kept for the next iterate. Notably, the final time $T$ (and the final number of snapshots) does not need to be known ahead of time as the snapshots are briefly held in memory to adjust the weights accordingly.

It is possible to update the integral a single snapshot at a time in order to cut memory costs if the final snapshot is signaled $P$ snapshots ahead to account for final adjustments, i.e. either complete the final set of $P$ many snapshots or adjust to a lower order Newton-Cotes. In this case we have, 
\begin{equation}
\label{eq:int_scheme2}
I(n+1) = I(n)  +\alpha_P \cdot \Delta t \cdot w_{p(n+1)} g_{n+1} \quad \text{ where } p(n+1)\in \{1, \ldots, P\}. 
\end{equation}
where $n$ is the number of the snapshot and $w_{p(n+1)}$ is the adjusted weight for the $n+1$-th snapshot. Using the composite trapezoid rule, 
\begin{equation}
    \int_0^T g(t) \,dt \approx \frac{\Delta t}{2} \cdot ( g(t_1) + 2\cdot g(t_2) + \ldots +  2\cdot g(t_{N-1}) + g(t_N)),
\end{equation}
is efficient in the sense that it does not require either holding multiple snapshots in memory or an early end signal as the final snapshot's weight can be adjusted by scaling. It is also possible to do streaming integration for snapshots taken at irregular intervals by adapting \cite{irregulart_integration}. In this case, the widths of the sub-intervals need to be recorded for the computation. Finally, as discussed in Section \ref{subsec:weak-SINDy}, many possible choices exist for both basis and test functions. The proposed streaming weak-SINDy approach could accommodate various basis and test functions, as well as some numerical integration schemes other than the Newton-Cotes formula. As shown in \cite{Gurevich}, proper choices of basis and test functions could lead to better accuracy.  

\subsection{Description of streaming weak-SINDy algorithm for compression}
\label{subsec:streaming_weak-SINDy_alg}

In this section, we describe the streaming weak-SINDy algorithm for streaming data compression.
To show that the algorithm can be performed in an online fashion, we start by rewriting the target vector $\bm{b}$ and the feature matrix $\bm{G}$ as
\begin{equation}
\label{eq:target_vector_integral}
\bm{b}=\int_0^T -\hat{\bm{b}}(t)\,dt:=
\int_0^T -\left [\bm{u}(t) \dot{\psi}_1(t), \ldots, \bm{u}(t) \dot{\psi}_K(t)\right]^\top\, dt 
\end{equation}
and
\begin{equation}
\label{eq:feature_matrix_integral}
\bm{G}= \int_0^T \hat{\bm{G}}(t)\, dt:=\bigintss_0^T \begin{bmatrix}\varphi_{1}(\bm{u}(t)) \psi_{1}(t) & \ldots &  \varphi_{J}(\bm{u}(t)) \psi_{1}(t)\\
\vdots & \ddots & \vdots \\
\varphi_{1}(\bm{u}(t)) \psi_{K}(t) & \ldots &  \varphi_{J}(\bm{u}(t)) \psi_{K}(t)
\end{bmatrix}\, dt, 
\end{equation}
respectively, where $\hat{\bm{b}}(t)$ and $\hat{\bm{G}}(t)$ are defined in the obvious way. 
In Algorithm~\ref{alg:streaming_weak_SINDy}, we describe the procedures of the basic streaming weak-SINDy algorithm with the details of streaming integration given in Function~\ref{func:UpdateFunction}.
Here, the test functions $\{\psi_k(t)\}_{k=1}^K$ are not assumed to have compact supports in $[0,T]$, which results in the initial and final terms in the target vector in Lines 1 and 5 in Algorithm~\ref{alg:streaming_weak_SINDy}, respectively. 
Algorithm~\ref{alg:streaming_weak_SINDy} shows that the online stage does not require storage of the snapshot data $\bm{u}(t_n)$ and that the sizes of $\bm{G}$ and $\bm{b}$ are constant over time, which make this streaming weak-SINDy approach memory-efficient, as analyzed below.
 
\paragraph{Efficiency:} Overall, the size of the data set is given by $\mu(\{\bm{u}_n\}) = SN$. At the same time, the size of all target vectors is given by $\mu(\{\bm{b}\}^S_{s=1})= SK$ and the size of the feature matrix is given by $\mu(\bm{G}) = JK$. As shown in Algorithm~\ref{alg:streaming_weak_SINDy} below, the target vectors and feature matrix are updated as more data becomes available in the online stage. This online compression process is efficient when $K(J+S)< SN$, which implies that the total online storage requirement is less than the requirement for storing the entire data set. 

After the data is streamed, the regression problems defined by the feature matrix and target vectors are solved in an offline stage, which gives a set of coefficients $\{\bm{c}_s\}^S_{s=1}$ with $\mu(\{\bm{c}_s\}) = SJ$. The target vectors and feature matrix can then be abandoned, since the data can be reconstructed through solving the surrogate model given by $\{\bm{c}_s\}^S_{s=1}$. The overall compression algorithm is then successful if  $SJ < SN$, but it is typical to have $J\ll N$.

\begin{Function}
  \SetKwFunction{FUpdate}{Update}
  \SetKwProg{Pn}{Function}{:}{}
  \Pn{\FUpdate{$t_n, \bm{u}(t_n)$, $\{\varphi_j\}^J_{j=1}$,$\{\psi_k\}^K_{k=1}$, $\bm{b}$, $\bm{G}$}}{

            Compute $\hat{\bm{b}}(t_n) = [\bm{u}(t_n)\dot{\psi}_1(t_n), \ldots, \bm{u}(t_n) \dot{\psi}_K(t_n)]^\top$\\

            Compute $\hat{\bm{G}}(t_n) = \left[\varphi_{j}(\bm{u}(t_n)) \psi_{k}(t_n)\right]^{K,J}_{k,j=1,1}$\\

            Compute 
            \begin{equation}
                \label{eq:b_update}
                \bm{b}_{\text{update}} = 
                -\hat{\bm{b}}(t_n)\cdot \alpha_{P}\cdot \Delta t \cdot w_{p(n)}        
            \end{equation} 
            
            Compute
            \begin{equation}
                \label{eq:G_update}
                \bm{G}_{\text{update}} = \hat{\bm{G}}(t_n)\cdot \alpha_{P}\cdot \Delta t \cdot w_{p(n)}       
            \end{equation}\\
            Update $\bm{b}\leftarrow \bm{b}+\bm{b}_{\text{update}} $\\
            Update $\bm{G}\leftarrow \bm{G}+\bm{G}_{\text{update}} $\\

    \KwRet $\bm{b}, \bm{G}$

  }
\caption{Update function}
\label{func:UpdateFunction}
\end{Function}

\begin{algorithm2e}
 
 \KwInitialize{Projection basis $\{\varphi_j(x)\}_{j=1}^J$, test basis $\{\psi_k\}_{k=1}^K$, $\bm{G} = \bm{0}\in \mathbb{R}^{K\times J}$, $\bm{b} = \bm{0}\in \mathbb{R}^{K}$ .}
\nonl\textit{Online:}\\
    $\bm{b} \leftarrow -[\bm{u}(t_1){\psi}_1(t_1), \ldots, \bm{u}(t_1){\psi}_K(t_1)]^\top$

    \For{$n\in\{1,2, \ldots N\}$}{

    $\bm{b},\bm{G}=\texttt{Update}(t_n, \bm{u}(t_n), \{\varphi_j\}^J_{j=1}, \{\psi_k\}^K_{k=1}, \bm{b}, \bm{G})$\\
    }
$\bm{b} \leftarrow\bm{b} + [\bm{u}(t_N){\psi}_1(t_N), \ldots, \bm{u}(t_N){\psi}_K(t_N)]^\top$

\tcc{The total number of snapshots $N$ is not required to start the algorithm. The step in Line~5 can be performed whenever the data stream ends.}  

\nonl\textit{Offline:}\\
    Compute solution $\bm{c}^*$ to $\bm{b} = \bm{G}\bm{c}$ via regression method of choice.

\Return{$\bm{c}^*$}
 \caption{Streaming weak-SINDy}
 \label{alg:streaming_weak_SINDy}
\end{algorithm2e}

\section{High dimensional data compression}
\label{sec:high-dimensional_data_compression}
In Section~\ref{sec:Streaming_weak_SINDy_for_compression}, we have shown that a streaming weak-SINDy model can compress a data stream $\{\bm{u}_n\}_{n=1}^N$ by using the fact that the target vectors and feature matrices are often smaller than the data set itself. In particular, we have that $\mu(\{\bm{b}_{s}\}\cup \{\bm{G}\}) = SK + JK < \mu(\{\bm{u}_n\}) = SN$ with $N$ being the number of time steps and often much larger in comparison to $S$,$J$, and $K$. Here, $J$ and $K$ are user chosen parameters as they are the sizes of the test and projection function bases. However, some data sets have very large state space dimension $S$, as is often the case for discretized PDE solutions. For a PDE solution $u:\Omega\times[0,T]\rightarrow \mathbb{R}$, the (compact) spatial domain $\Omega$ is discretized as $\hat{\bm{x}} = [x_1, \ldots, x_S]$ and a snapshot $u$ is of the form $\bm{u}(\hat{\bm{x}}, t_n) = [u(x_1, t_n), \ldots u(x_S,t_n)]^\top\in \mathbb{R}^S$, where $S$ is often large. 

The large state dimension $S$ impacts the weak-SINDy compression scheme in two ways. Foremost, the size of the set of target vectors, $\mu(\{\bm{b}_s\}) = SK$, is obviously dependent on $S$. Secondly, given a desirable projection accuracy, the cardinality of the projection basis, $J$, typically grows with the size of the state space $S$. For instance, consider a monomial basis $u_1^{r_1}u_2^{r_2}\cdot \ldots \cdot u_S^{r_S}$ of max degree $R$, we have $(R+1)^S$ many functions. Even when considering polynomials of total degree $R$, for large $S$ and small $R$, the number of monomial basis functions is $O(S^R)$. However, when considering surrogate modeling, \cite{russo2022convergence} shows that if monomials are chosen as a projection basis for weak-SINDy, the reconstruction accuracy of the surrogate model increases with the max degree of the polynomial model. Hence, there is a balance between the size of the basis, which impacts the compression, and the expressivity of the basis, which affects the accuracy. 

Hence, the impacts of a large state dimension $S$ merits the use of a dimension reduction technique for high state dimension problems. In this work, we reduce the state dimension using a streaming POD approach and apply the streaming weak-SINDy algorithm to compress the temporal POD modes. Although combinations of POD and regression-based surrogate modeling techniques have been explored in the literature \cite{liftandlearn,russo2022convergence,PhysRevE.104.015206}, they are not directly applicable to the streaming setting considered here. 
In this section, we extend the streaming weak-SINDy algorithm to accommodate a POD basis of varying size from a streaming POD method. The resulting algorithm provides a memory efficient approach to compress streaming high dimensional data.

Section~\ref{subsec:generating_surrogate_ODE_for_POD} is dedicated to the integration of the POD dimension reduction technique into the streaming weak-SINDy algorithm. There we assume the data ${D} = \{\bm{u}_n\}$ at each time $t_n$ is given as a sample of a solution $u(\bm{x},t)$ to a partial differential equation $\partial_t u(\bm{x},t) = F(u(\bm{x},t))$. 
Section~\ref{subsec:streaming_pod} introduces a streaming POD method, adapted from \cite{streaming_POD}, which constructs POD modes from streaming data and updates the modes in an online manner. 
Sections~\ref{subsec:generating_surrogate_ODE_for_POD} and \ref{subsec:streaming_pod} essentially serve as background information. The main contribution of this work is presented in Section~\ref{subsec:modify_weak_SINDy}, which details the approach we take to integrate the streaming POD method to the streaming weak-SINDy algorithm and analyzes the memory cost of the resulting algorithm.

\subsection{Generating a surrogate ODE system for the POD modes}
\label{subsec:generating_surrogate_ODE_for_POD}

With the POD modes defined in Section~\ref{subsec:POD}, the POD expansion of data $u$ can written as $u(\bm{x},t) = \sum_{\ell=1}^\infty \nu_\ell(t)\upsilon_\ell(\bm{x})$. Assuming that the evolution of the first $L$ temporal modes is governed by an ODE system with some dynamics $f\colon\mathbb{R}^L\to\mathbb{R}^L$, i.e.,
\begin{equation}\label{eq:exact_POD_system}
\dot{\bm{\nu}}(t)= f(\bm{\nu}(t))\:
\quad\text{with}\quad
\bm{\nu}(t):=\begin{bmatrix}\nu_1(t)& \cdots & \nu_L(t)\end{bmatrix}^T\:,
\end{equation}
then, the weak-SINDy method can be applied to construct a surrogate model for $f$ from data $\bm{\nu}$. Constructing an approximation of $f$ allows us to model the temporal modes $\nu_\ell(t)$, while the spatial modes $\upsilon_\ell(x)$ remain constant. Applying the weak-SINDy method to Eq.~\eqref{eq:exact_POD_system} leads to the surrogate model
\begin{equation}\label{eq:PODmodel}
\textstyle
\dot{\nu}_\ell(t) = \sum_{{j}=1}^J\bm{c}_{j}^{(\ell)} \cdot \varphi_{j}(\bm{\nu}(t))\:,\quad \ell = 1,\dots, L,
\end{equation}
where the weights $\bm{c}_{j}^{(\ell)}$ are computed via regression. At this stage, we can now see the effect of this dimension reduction on the streaming weak-SINDy algorithm. If a streaming POD algorithm is injected as a dimension reduction step before the streaming weak-SINDy algorithm, the state dimension becomes $L\ll S$. Applied to the temporal POD mode data stream, we now have $\mu(\{\bm{b}_\ell\}^L_{\ell=1}\cup\bm{G}) = LK + JK$. If using a monomial projection basis, the cardinality of the basis, $J$, is now also dependent on $L$, which is much smaller than the original state dimension $S$. In the remainder of this paper, we refer to the solutions ${\tilde{\nu}_\ell}$ of Eq.~\eqref{eq:PODmodel} as the \emph{surrogate} temporal modes, which can be used to construct an approximation of the data $u(\bm{x},t)\approx \tilde{u}(\bm{x},t):=\sum_{\ell=1}^L \tilde{\nu}_\ell (t) \upsilon_\ell(\bm{x})$.

\subsection{Streaming POD}
\label{subsec:streaming_pod}

In an offline compression process, all snapshots of $u$ would be recorded. Upon resolving the spatial modes and temporal modes, unimportant modes, as indicated by the magnitude of corresponding singular values, can be discarded. However, this is not feasible in the streaming setting. Spatial POD modes can be generated in an online process by allowing for a small amount of working memory. For $p_{0}\in \mathbb{N}$, $1<p_{0}<N$, the first $p_{0}$ snapshots of the total $N$ snapshots are collected to form the matrix $\bm{D}_{p_{0}} = [\bm{u}(\hat{\bm{x}}, t_1), \ldots \bm{u}(\hat{\bm{x}},t_{p_{0}})]\in \mathbb{R}^{S\times p_{0}}$. The spatial modes $\bm{U}_{p_{0}}$ are computed via the snapshot method described in Section~\ref{subsec:POD}, and $\bm{U}^L_{p_{0}}$ is built from the first $L$ columns of $\bm{U}_{p_{0}}$. 

\begin{figure}
\centering
\includegraphics[width = \textwidth]{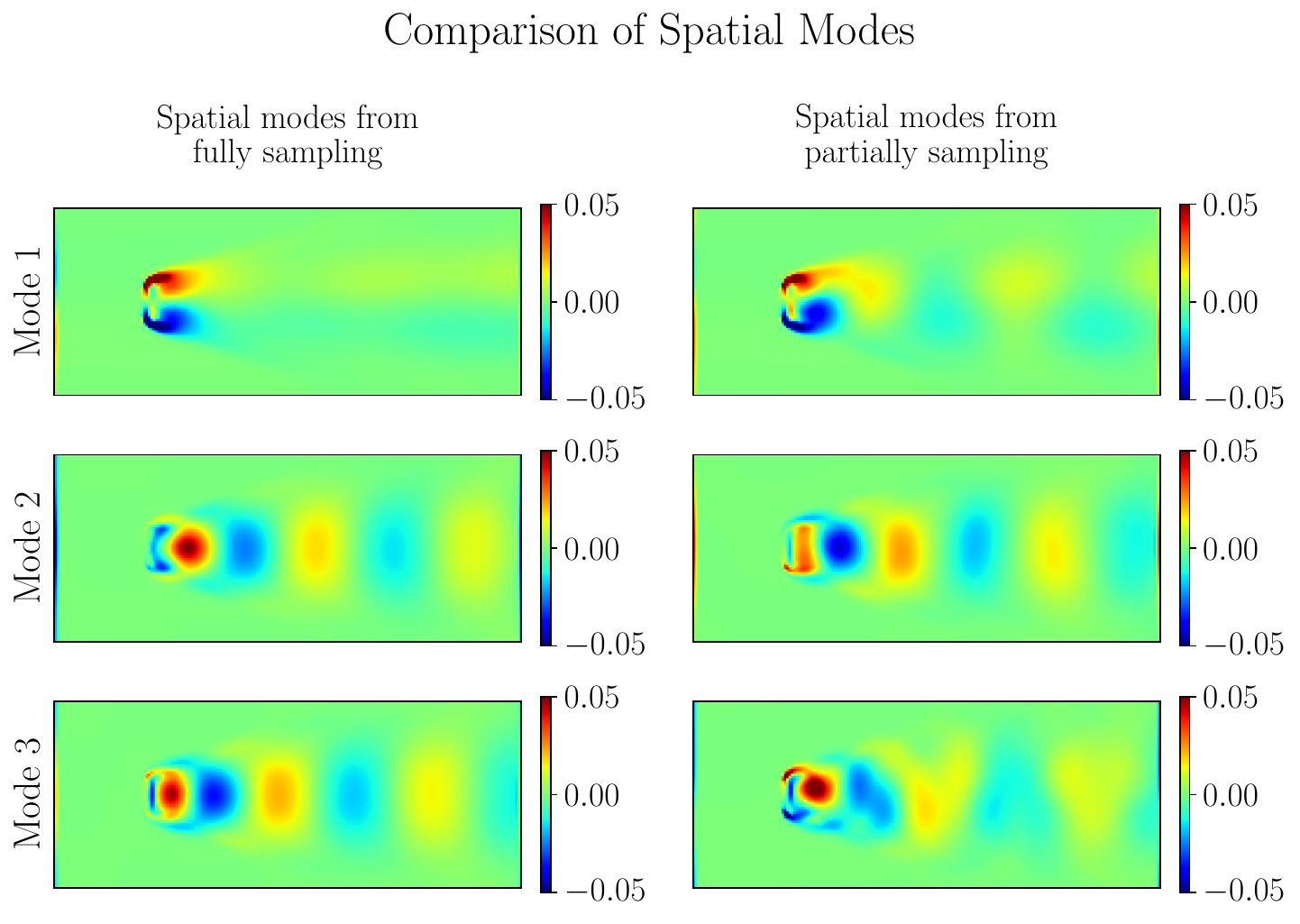}
\caption{A comparison of the first three spatial modes of an example data set. The left panel shows the spatial modes constructed using the entire set of $10,000$ snapshots. The right panel shows spatial modes constructed from using a partial sampling of the first $550$ snapshots.}
\label{fig:fullysampled_vs_subsampled_POD_modes}
\end{figure}

As indicated by Eq.~\eqref{eq:R-formula}, the POD modes capture the average spatial behavior over the time interval $[t_1,t_{p_{0}}]$. Since only the first ${p_{0}}$ many snapshots of the total number of snapshots $N$ are used, it is unlikely that the resulting initial POD modes $\bm{U}_{p_{0}}$ gives a good representation to all $N$ snapshots (see Figure~\ref{fig:fullysampled_vs_subsampled_POD_modes} for an example). Hence, in Function~\ref{func:streaming_POD}, we describe a streaming POD model updating technique that is applied to each snapshot. This function either uses the snapshot to build an initial basis (as described above) or evaluates the current basis and adds in new elements when needed. The application of Function \ref{func:streaming_POD} to each snapshot will be referred to as ``streaming POD". This is adapted from \cite{streaming_POD} for its computational simplicity and its ability to capture new information that is not accurately expressible in terms of the current POD modes. 

\begin{Function}
\SetKwProg{Pn}{Function}{:}{}
\SetKwFunction{FPOD}{UpdatePOD}
\Pn{\FPOD{$\varepsilon_{\text{spec}}$, $\varepsilon_{\text{res}}$, $p_{0}$, $\bm{u}(\hat{\bm{x}},t_n)$}}{
    {$\bm{D}_{0}=\emptyset$\;}
    \If{$n$ in $\{1,2, \ldots, p_{0}\}$}
        {
        {$\bm{D}_{n} \leftarrow [\bm{D}_{n-1}, \bm{u}(\hat{\bm{x}}, t_n)]$}\tcp*{Record $\bm{u}(\hat{\bm{x}}, t_n)$}
        \Return{$\emptyset$}
        }
    \If{$n=p_0$}{
        Compute $[\bm{U}_{p_{0}}, \bm{\Sigma}_{p_{0}}, \bm{V}_{p_{0}}] = \operatorname{svd}(\bm{D}_{p_{0}})$\;
        Let $L$ be the largest index for which $\sigma_L\geq \varepsilon_{\text{spec}}$, where $\sigma_L$ is a diagonal entry of $\bm{\Sigma}_{p_{0}}$.\;
        $\bm{P}\leftarrow\bm{U}^L_{p_{0}}$\tcp*{Set initial spatial modes}
        \Return{$\bm{P}$, $\bm{\Sigma}^L_{p_{0}}$, $\bm{V}^L_{p_{0}}$}
    }
    \If{$n$ in $\{p_0+1, \ldots, N\}$}{
            $r \leftarrow \operatorname{res}(\bm{u}(\hat{\bm{x}},t_n),\bm{P})$\tcp*{Compute residual for current spatial modes}
                \If{$r>\varepsilon_{\text{res}}$}{
                    $\bm{p}_{\perp} \leftarrow \left(\textbf{Id} - \bm{P} \cdot \bm{P}^\top\right)\cdot \bm{u}(\hat{\bm{x}},t_n)$\;
                    $\bm{P}\leftarrow \left[\bm{P}, \frac{\bm{p}_{\perp}}{\|\bm{p}_{\perp}\|_2}\right]$\tcp*{Add a new spatial mode}
                }
            \Return{$\bm{P}$}
            }

    }

 \caption{Streaming POD}
 \label{func:streaming_POD}
\end{Function}

For a general orthogonal matrix $\bm{U}$ and vector $\bm{v}$, we define $\operatorname{res}(\bm{v}, \bm{U})$, the residual of $\bm{v}$ with respect to $\bm{U}$ used in Function~\ref{func:streaming_POD} as 
\begin{equation}
\label{eq:residual}
\operatorname{res}(\bm{v}, \bm{U}) = \|\left(\textbf{Id} - \bm{U} \cdot \bm{U}^\top\right) \cdot \bm{v}\|_2 / \|\bm{v}\|_2,
\end{equation} 
where $\textbf{Id}$ is the identity matrix with appropriate size. Starting from the initial orthogonal spatial POD modes $\bm{U}^L_{p_{0}}$, we compute the residual $\operatorname{res}(\bm{u}(\hat{\bm{x}}, t_n), \bm{U}_{p_{0}}^L)$ at each incoming data $\bm{u}(\hat{\bm{x}}, t_n)$. When the residual exceeds a user-selected threshold, the vector $\left(\textbf{Id} - \bm{U}_{p_0}^L \cdot (\bm{U}_{p_{0}}^L)^\top\right) \cdot \bm{u}(\hat{\bm{x}}, t_n)/\|\left(\textbf{Id} - \bm{U}_{p_{0}}^L \cdot (\bm{U}_{p_{0}}^L)^\top\right) \cdot \bm{u}(\hat{\bm{x}}, t_n)\|_2$ is added in as a (discretized) spatial mode. From a functional point of view, the residual is approximating $\|u(\bm{x},t_n) - \operatorname{Proj}u(\bm{x},t_n)\|_\Omega/\|u(\bm{x},t_n)\|_\Omega$, where the projection is onto the subspace defined by the first $L$ spatial modes. 

\begin{notation}
In Function \ref{func:streaming_POD}, the notation $\bm{C} = [\bm{A},\bm{b}]\in\mathbb{R}^{p\times (q+1)}$ denotes the matrix from concatenating $\bm{A}\in\mathbb{R}^{p\times q}$ and $\bm{b}\in \mathbb{R}^p$. This notation is extended to denote concatenations of two matrices of appropriate sizes later in the article. 
\end{notation}

\subsection{Combining streaming weak-SINDy and streaming POD methods}
\label{subsec:modify_weak_SINDy}
Initially, the target vectors and feature matrix can be created using temporal data from the SVD procedure which generates the initial POD basis. This temporal data is contained in the matrices $\bm{\Sigma}^L_{p_0}$ and $\bm{V}^L_{p_0}$ (defined analogously to $\bm{U}^L_{p_0}$).  As the streaming POD method adds new spatial modes when the snapshot data $\bm{u}(\hat{\bm{x}}, t_n)$ are taken in, the streaming weak-SINDy method needs to incorporate the new POD modes. Suppose that for timestamps $\{t_1, \ldots, t_{\hat{n}}\}$ with $1<\hat{n}<N$, the feature matrix $\bm{G}$ and target vectors $\bm{b}$ are collected based on the first $L$ spatial POD modes $\{\upsilon_1, \ldots, \upsilon_L\}$. With $\bm{\nu}(t) = [\nu_1(t), \ldots, \nu_L(t)]^\top$, we have 
\begin{equation} 
\label{eq:bvect_Lmodes}
[\bm{b}_1, \ldots, \bm{b}_L]  = \left[\begin{array}{c:c:c:c}
\ip{\dot{\nu}_1}{\psi_1}_{[0,t_{\hat{n}}]} & \ip{\dot{\nu}_2}{\psi_1}_{[0,t_{\hat{n}}]}& \cdots & \ip{\dot{\nu}_L}{\psi_1}_{[0,t_{\hat{n}}]}\\
\vdots & \vdots & \cdots & \vdots\\
\ip{\dot{\nu}_1}{\psi_K}_{[0,t_{\hat{n}}]} & \ip{\dot{\nu}_2}{\psi_K}_{[0,t_{\hat{n}}]}& \cdots & \ip{\dot{\nu}_L}{\psi_K}_{[0,t_{\hat{n}}]}\\
\end{array}\right]
\end{equation}
and
\begin{equation}
\label{eq:G_Lmodes}
\bm{G} = \left[\begin{array}{c}
\ip{\varphi_j(\bm{\nu}(t))}{\psi_k(t)}_{[0,t_{\hat{n}}]}\\
\end{array}\right]^{K,J}_{k=1, j=1}.
\end{equation}
Now, suppose at time $t_{\hat{n}+1}$, a new spatial POD mode is added. We then modify the target vectors and feature matrix in the following way. Let $\tau$ denote either the time step in which the next new mode is added or the final time, and let $\nu_{L+1}$ denote the new temporal mode.
The modified target vector takes the form $[\bm{b}_1^*, \ldots, \bm{b}^*_L]$ with
\begin{equation}
\label{eq:bvect_L+1modes}
 \bm{b}^*_\ell = \left[\ip{\dot{\nu}_\ell}{\psi_1}_{[0,t_{\hat{n}}]},\ldots,\ip{\dot{\nu}_\ell}{\psi_K}_{[0,t_{\hat{n}}]},
 \ip{\dot{\nu}_\ell}{\psi_1}_{[t_{\hat{n}+1}, \tau]},\ldots
\ip{\dot{\nu}_\ell}{\psi_K}_{[t_{\hat{n}+1}, \tau]} \right]^\top
\end{equation}
and an additional target vector for the new temporal mode $\nu_{L+1}$:
\begin{equation}
\label{eq:new_target_vect}
\bm{b}_{L+1} = \big[\ip{\dot{\nu}_{L+1}}{\psi_1}_{[t_{\hat{n}+1}, \tau]}, \ldots, \ip{\dot{\nu}_{L+1}}{\psi_K}_{[t_{\hat{n}+1}, \tau]}\big]^\top.
\end{equation}
As a new mode $\nu_{L+1}(t)$ is being added, we necessarily expand the size of the basis $\{\varphi_{j}\}_{j=1}^{J}$ by $J'$ many more functions. Hence we modify $\bm{G}$ in the following fashion. Let $\bm{\nu}^* = [\nu_1(t), \ldots, \nu_{L}(t), \nu_{L+1}(t)]^\top$,
\begin{equation}
\label{eq:G_L+1modes}
\bm{G}^* = \begin{bmatrix}\bm{G}_1 & 0 \\ \bm{G}_2 & \bm{G}_3\end{bmatrix}
\quad
\bm{G}_{1,k,j} = 
\ip{\varphi_j(\bm{\nu}(t))}{\psi_k(t)}_{[0,t_{\hat{n}}]}
\quad \substack{j=1, \ldots, J\\ k = 1,\ldots, K}
\end{equation}
\begin{equation}
\bm{G}_{2,k,j} = \ip{\varphi_j(\bm{\nu}^*(t))}{\psi_k(t)}_{[t_{\hat{n}+1}, \tau]}\quad \substack{j=1, \ldots, J\\ k = 1,\ldots, K}
\end{equation}
\begin{equation}
\bm{G}_{3,k,j} = \ip{\varphi_j(\bm{\nu}^*(t))}{\psi_k(t)}_{[t_{\hat{n}+1}, \tau]} \quad \substack{j=J+1, \ldots, J'\\ k = 1,\ldots, K}
\end{equation}
Here, we make an assumption that the basis functions $\{\varphi_j\}$ are given by tensor product, i.e. $\varphi_j(\bm{\nu}(t)) = \varphi_j(\bm{\nu}^*(t))$
for $j=1,\ldots, J$, which is satisfied by mulitvariable monomials among with many other choices of basis functions. Now, in the offline stage, to build surrogate models for $\nu_\ell(t)$ with $\ell = 1,\ldots, L$, we solve 
\begin{equation}
\label{eq:solve_first_L}
\bm{G}^*\bm{c} = \bm{b}^*_\ell. 
\end{equation}
As for building a surrogate model for $\nu_{L+1}(t)$, we solve
\begin{equation}
\label{eq:solve_L+1}
[\bm{G}_2,\, \bm{G}_3]\bm{c} = \bm{b}_{L+1}. 
\end{equation}
The process can be repeatedly applied when the addition of new modes occurs at multiple time steps as summarized in the following subsection. 

\subsubsection{Algorithm for modifying weak-SINDy}
  In this section we summarize the above development as a function (Function \ref{func:Build_and_Solve} -- \texttt{BuildAndSolve}) and give the pseudocode for the combination of the streaming POD mode update function (Function \ref{func:streaming_POD} -- \texttt{UpdatePOD}) and the streaming weak-SINDy method (Algorithm \ref{alg:streaming_weak_SINDy}) in Algorithm \ref{alg:weak-PSINDy}. In particular, Algorithm \ref{alg:weak-PSINDy} details the creation and storage of feature matrices and target vectors needed to create the (generalized) expanded forms in Eqs.~\eqref{eq:new_target_vect} and \eqref{eq:G_L+1modes}. 

\begin{algorithm2e}
 
 \KwInitialize{Projection basis $\{\varphi_j(x)\}_{j=1}^J$, test basis $\{\psi_k\}_{k=1}^K$, $\operatorname{Modes}=\{\ \}$, $\operatorname{Problem}=\{\ \}$, $\hat{L} = 0$, $M = 0$, $J_0=J$.}
\nonl\ \\
\nonl\textit{Online:}\\
    \For{$n$ in $\{1,2, \ldots, p_0-1\}$}{
        $\emptyset = \texttt{UpdatePOD}(\varepsilon_{\text{spec}}, \varepsilon_{\text{res}}, p_0,\bm{u}(\hat{\bm{x}},t_n))$\tcp*{Initial data collection}
    }
    $[\bm{P}, \bm{\Sigma}^L_{p_0}, \bm{V}^L_{p_0}] = \texttt{UpdatePOD}(\varepsilon_{\text{spec}}, \varepsilon_{\text{res}}, p_0,\bm{u}(\hat{\bm{x}},t_{p_0}))$\tcp*{Compute initial POD modes}
    Compute $\{[\nu_\ell(t_1), \ldots \nu_\ell(t_{p_0})]^\top \mid \ell = 1, \ldots L\}$ ( = $\bm{\Sigma}^L_{p_0}\cdot \bm{V}^L_{p_0}$ ) \;

    Use $\{[\nu_\ell(t_1), \ldots \nu_\ell(t_{p_0})]^\top \mid \ell = 1, \ldots L\}$ to initialize $\bm{b} = [\bm{b}_1, \ldots, \bm{b}_L]$ and $\bm{G}$ (see Equation~\eqref{eq:weak-SINDy_form})\;

    $\hat{L}\leftarrow L$\;
    $\operatorname{Modes} \leftarrow \bm{P} (= \{\bm{\upsilon}_\ell(x)\, | \,\ell = 1,\ldots L\} )$\;

    \For{$n$ in $\{p_0+1, \ldots, N\}$}{

    Compute $\{\bm{\upsilon}_\ell(x)\, | \,\ell = 1,\ldots L\} = \texttt{UpdatePOD}(\varepsilon_{\text{spec}}, \varepsilon_{\text{res}}, p_0,\bm{u}(\hat{\bm{x}},t_n))$\tcp*{Update POD modes}
    
\tcc{When a new mode is added, store the feature matrix and target vectors for current modes and expand the basis for the new mode.}
     \If{$|\{\bm{\upsilon}_\ell(x)\, | \,\ell = 1,\ldots L\}|>\hat{L}$}{
        
        $\operatorname{Modes} \leftarrow \{\bm{\upsilon}_\ell(x)\, | \,\ell = 1,\ldots L\}$\;

        $\bm{b} \leftarrow \bm{b} + [\bm{\nu}(t_{n-1}) {\psi}_1(t_{n-1}), \ldots, \bm{\nu}(t_{n-1}) {\psi}_K(t_{n-1})]^\top$\;
        
        $(\bm{b}^{(M)}, \bm{G}^{(M)}) \leftarrow (\bm{b}, \bm{G})$\;

        $\operatorname{Problem} \leftarrow \operatorname{Problem}\cup \{(\bm{b}^{(M)},\bm{G}^{(M)})\}$\;

        $M\leftarrow M+1$\;

        Expand projection basis $\{\varphi_j\}^{J_{M-1}}_{j=1}\leftarrow \{\varphi_j\}^{J_M}_{j=1}$ with $J_M>J_{M-1}$ \;
    }
    Compute $\{\nu_\ell(t_n) = \ip{u(\bm{x},t_n)}{\upsilon_\ell(\bm{x})}_\Omega \, | \, \ell = 1, \ldots L\}$\tcp*{Compute temporal modes}
\tcc{Initialize $(\bm{b}, \bm{G})$ when a new mode is added.}
    \If{$|\{\bm{\nu}_\ell(t_n)\, | \,\ell = 1,\ldots L\}|>\hat{L}$}{

        $(\bm{b}, \bm{G}) \leftarrow (-[\bm{\nu}(t_{n})\cdot {\psi}_1(t_{n}), \ldots, \bm{\nu}(t_{n})\cdot {\psi}_K(t_{n})]^\top, \bm{0})$\;
        
        $\hat{L} \leftarrow \hat{L}+1$\;
    }

    Compute $\bm{b},\bm{G} = \texttt{Update}(t_n, \bm{\nu}(t_n),\{\varphi_j\}^{J}_{j=1}, \{\psi_k\}^K_{k=1}, \bm{b}, \bm{G})$\tcp*{Update $(\bm{b}, \bm{G})$}

    }

\tcc{Store $(\bm{b}, \bm{G})$ when data stream finishes.}
$\bm{b} \leftarrow \bm{b} + [\bm{\nu}(t_{N}) {\psi}_1(t_{N}), \ldots, \bm{\nu}(t_{N}) {\psi}_K(t_{N})]^\top$\;

$(\bm{b}^{(M)}, \bm{G}^{(M)}) \leftarrow (\bm{b}, \bm{G})$\;

$\operatorname{Problem} \leftarrow \operatorname{Problem}\cup \{(\bm{b}^{(M)},\bm{G}^{(M)})\}$\;
    
\nonl\ \\
\nonl\textit{Offline:}\\
    Compute $\bm{c^*} = \texttt{BuildAndSolve}(\operatorname{Problem})$\;
    Discard $\operatorname{Problem}$\;
\nonl\ \\
\Return{$\bm{c}^*$, $\operatorname{Modes}$}
 \caption{Streaming weak-SINDy and POD}
 \label{alg:weak-PSINDy}
\end{algorithm2e}

With Algorithm \ref{alg:weak-PSINDy} detailed, we can now outline the regression procedure called $\texttt{BuildAndSolve}$ referenced in Algorithm \ref{alg:weak-PSINDy}. Suppose $M$ new spatial modes are added at times $t_{n_1}< \ldots< t_{n_M}$. Each time a new mode is added, the projection basis functions is updated to $J_m$ many functions and the temporal modes are updated as well. For convenience, we define $L_m:=L+m$. Now, since $M$ new modes are added, this means the ``Problem" set in Algorithm \ref{alg:weak-PSINDy} has been appended to $M+1$ times.
In Algorithm \ref{alg:weak-PSINDy}, the appended $(\bm{b}, \bm{G})$ are indexed by the order in which they are appended. The pairs of $(\bm{b},\bm{G})$ appended to the Problem set are again denoted by $\left([\bm{b}^{(m)}_1,\ldots, \bm{b}^{(m)}_{L_m}], \bm{G}^{(m)}\right)$.  Here, we have added the sub-index to $\bm{b}^{(m)}$ to indicate the number of data vectors. Note that $\bm{G}^{(m)}$ can be decomposed as 
\begin{equation}
 \bm{G}^{(m)}=[\hat{\bm{G}}^{(m)}, \tilde{\bm{G}}^{(m)}] 
\end{equation}
where $\tilde{\bm{G}}^{(m)}$ are the additional $J_m-J_{m-1}$-many columns that $\bm{G}^{(m)}$ has in comparison to $\bm{G}^{(m-1)}$ as this corresponds to an expanded projection basis. 

\begin{Function}
\SetKwProg{Pn}{Function}{:}{}
\SetKwFunction{FBuildandSolve}{BuildAndSolve}
\Pn{\FBuildandSolve{$\left\{\left([\bm{b}^{(m)}_1,\ldots, \bm{b}^{({m})}_{L_m}], \bm{G}^{(m)}\right)\, \mid \, m = 1,\ldots, M\right\}$}}{
\nonl \ \\
\nonl\textit{Build:}
\nonl\ \\
$\mathfrak{G}_0 = \bm{G}^{(0)}$\;
$[\mathfrak{b}_1, \ldots \mathfrak{b}_L] = [\bm{b}^{(0)}_1, \ldots, \bm{b}^{(0)}_L] $\;
    \For{$m$ in $\{1,2, \ldots, M\}$}
        {
        $\mathfrak{G}_m\leftarrow \left[\begin{array}{cc}
                \mathfrak{G}_{m-1} & \bm{0}\\
                \hat{\bm{G}}^{(m)}&\tilde{\bm{G}}^{(m)}\\
            \end{array}\right] $\;

            $[\mathfrak{b}_1, \ldots \mathfrak{b}_{L_{m-1}}, \mathfrak{b}_{L_m}] \leftarrow \left[
                \left[\begin{array}{c}
                        \mathfrak{b}_1\\
                        \ \\
                        \bm{b}^{(m)}_{1}
                \end{array}\right],
                \cdots,
                \left[\begin{array}{c}
                        \mathfrak{b}_{L_{m-1}}\\
                        \ \\
                       \bm{b}^{(m)}_{L_{m-1}}, 
                \end{array}\right], 
                 \bm{b}^{(m)}_{L_{m}}
            \right] 
        $\;
        }
    Relabel:
    $\mathfrak{G}_M = \left[\begin{array}{cccc}\mathcal{G}^\top_{L_0}\in \mathbb{R}^{J_M\times K} &
                    \mathcal{G}^\top_{L_1}\in \mathbb{R}^{J_M\times K}&
                    \cdots &
                    \mathcal{G}^\top_{L_M}\in \mathbb{R}^{J_M\times K}
                \end{array}\right]^\top 
    $\;
    \tcc{The matrix above is split along the rows into $M+1$ many matrices with $k$ rows each. }
\nonl\textit{Solve:}\\
        { Compute coefficient vectors using equations,
            \begin{equation}\mathfrak{b}_\ell = \mathfrak{G}_{M} \bm{c}_\ell \quad \text{ for }\quad \ell = 1,\ldots, L
            \end{equation}
            \begin{equation}\mathfrak{b}_{L_m} = [\mathcal{G}^\top_{L_m}, \ldots \mathcal{G}^\top_{L_M}]^\top \bm{c}_{L_m} \quad \text{ for }\quad m = 1,\ldots, M
            \end{equation}
            
        and regression method of choice. \\
        }
    \Return{$\bm{c}_1, \ldots,\bm{c}_L, \bm{c}_{L_1}, \ldots,\bm{c}_{L_M}$}

}
 \caption{Build and Solve}
 \label{func:Build_and_Solve}
\end{Function}

\begin{figure}[htp]
\begin{center}
\scalebox{.62}{%
\includegraphics[]{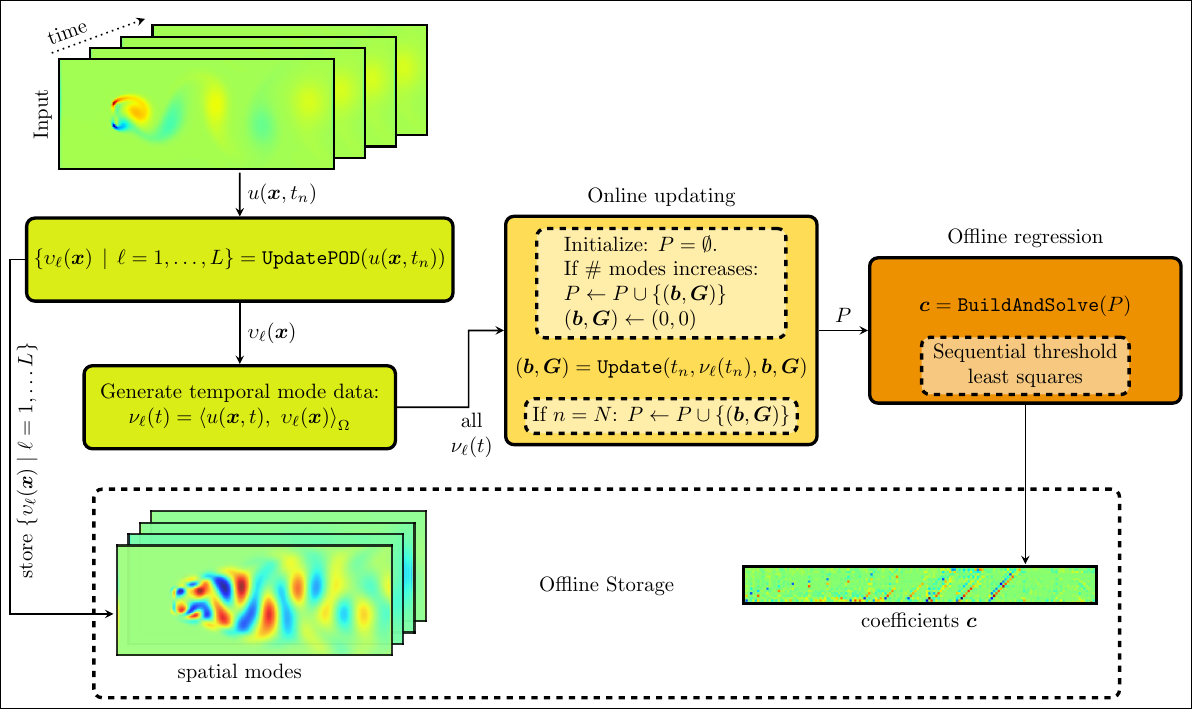}
}
\end{center}
\caption{This flowchart outlines the key steps in Algorithm \ref{alg:weak-PSINDy} and gives a visual illustration of the offline storage requirement of the algorithm.}
\label{fig:all_combined_algs}
\end{figure}

It should be noted that, Function \ref{func:Build_and_Solve} effectively replaces the offline portion of Algorithm \ref{alg:streaming_weak_SINDy}. Moreover, Algorithm \ref{alg:weak-PSINDy} temporarily stores the output of Function \ref{func:UpdateFunction} (\texttt{Update}) at points and this does increase the cost of the online portion as the feature matrix $\bm{G}_M$ and target vectors $[\bm{b}_1, \ldots , \bm{b}_{L_m}]$ are now larger in size. However, $\bm{G}_{M}$ is a lower triangular block matrix as shown by Eq.~\eqref{eq:G_L+1modes}. To analyze the cost, note $J<J_1<\ldots<J_M$ and that $\mathcal{G}_{1,\ldots,L}$ is of size $K\times J$ each $\mathcal{G}_{L_m}$ is of size $K\times J_m$. Therefore $\mu(\bm{G}_M) = K(J+J_1+\ldots + J_M)$. We also have $\mu(\{\bm{b}_1\ldots \bm{b}_L\}) = L(M+1)K$ and $\mu(\bm{b}_{L_m}) = (M+1-m)K$. Therefore $\mu(\{\bm{b}_{L_1}, \ldots, \bm{b}_{L_M}\}) = \frac{1}{2} (M^2+M)K$. It is clear that the growth of the matrices is dependent on $M$, the number of times a new mode is added. Hence, to keep this process efficient, it is best to minimize the number of times a new mode is added. Figure \ref{fig:all_combined_algs} describes Algorithm \ref{alg:weak-PSINDy}.

\subsubsection{Analysis of Algorithm \ref{alg:weak-PSINDy}}
\label{subsubsec:analysis_of_alg}

There are several factors that can make the online storage requirements of Algorithm \ref{alg:weak-PSINDy} less efficient. The addition of Algorithm \ref{alg:streaming_weak_SINDy} to the streaming POD process potentially compresses temporal mode data during the online process. The biggest storage requirement of Algorithm \ref{alg:weak-PSINDy} is the feature matrix $\bm{G}$. Hence, if the number of entries of $\bm{G}$ grows larger than the temporal mode data, the streaming weak-SINDy process becomes inefficient. The major contributing factors to the size of $\bm{G}$ is the number of projection functions $\{\varphi_j\}$ and the number of times new modes are added. Every time a new mode is added, it contributes a new row of blocks to the block form of $\bm{G}$ seen in Table \ref{table:form_feature_matrix_data_vec} and also requires new functions to be added to the set of projection functions. The overall size of the feature matrices and target vectors as discussed in Section \ref{subsec:modify_weak_SINDy} is summarized in Table \ref{table:size_formulas}.

\begin{table}[htp]
\centering
\begin{tabular}{l|c}
\label{table:size_formulats}
& \textbf{Size} \\
\hline
Feature matrix &  $K(J+J_1+\ldots + J_M)$ \\
\hline
Target vectors & $KL(M+1) + \frac{1}{2}(M^2+M)K$
\end{tabular}
\caption{Size of the target vectors and feature matrix after an adding $M$ additional modes. Here, $J$ is the original number of projection functions and each $J_m$ is the new number of projection functions after a new mode is added. Note, $J<J_1<\ldots<J_M$. }
\label{table:size_formulas}
\end{table}

An initial collection of snapshots is used to generate the initial spatial POD modes. As Figure \ref{fig:fullysampled_vs_subsampled_POD_modes} shows, in general the spatial modes generated from this initial sampling are different from the ones generated with more snapshots. From a functional point of view, the use of this initial collection of snapshots corresponds to a operator kernel 
\[R_\tau(\bm{x},\bm{y}) = \frac{1}{\tau}\int_0^\tau u(\bm{x}, s) u(\bm{y}, s) \, ds\]
which converges to the operator kernel given in Eq.~\eqref{eq:R-formula} as $\tau\rightarrow T$. Hence, a bigger collection of initial snapshots can potentially alleviate the need to add spatial modes in the streaming POD process. However, this comes at the cost of more online storage. Additionally, as stated in the introduction of Section \ref{sec:high-dimensional_data_compression}, an increase in the state dimension (in this case, adding additional POD modes), can increase the size of the basis dramatically. Hence, its necessary to selectively add basis functions. In Section \ref{subsec:Fluid_Flow}, the monomials were restricted to total degree 2, keeping the basis relatively small. Finally, this approach is not effective when the data rapidly changes as we will see in Section \ref{subsubsec:Deficiency} or when the data is not sufficiently regular as shown by Theorem \ref{thm:ChangandHa}.

\section{Numerical experiments}
\label{sec:Experiments}
In this section, we provide some proof of concept examples by applying Algorithm \ref{alg:streaming_weak_SINDy} to Lorenz system data as a low state dimension example and applying Algorithm~\ref{alg:weak-PSINDy} to a fluid flow data set as a high state dimension example. Here the ODEs, including the Lorentz system and the models from weak-SINDy, are evolved using the {\tt{odeint}} solver from the {\tt{scipy}} Python package. All non streaming integrals were either calculated using the {\tt{simpson}} or {\tt{trapezoid}} functions also found in the {\tt{scipy}} package. All streaming integrals were calculated using a streaming trapezoid rule. In the application of weak-SINDy, monomials were used as a projection basis and a Fourier basis $\{\psi_k\}^{2\tilde{K}+1}_{k=1} = \{ \sqrt{2/T} \sin(2\pi(k/T)t)\}^{\tilde{K}}_{k=1}\cup \{ \sqrt{2/T} \cos(2\pi(k/T)t)\}^{\tilde{K}}_{k=1}\cup \{1/\sqrt{T}\}$ was used as a test function basis. Here, $T$ is the length of the interval. Moreover, when referring to the degree of a monomial, total degree is defined as the sum of the powers and max degree is the largest power appearing. For clarity, we will call all models generated through an application of the standard weak-SINDy (Section \ref{subsec:weak-SINDy}) to the entire data set ``static'' models and the models generated from streaming weak-SINDy (Section \ref{subsec:streaming_weak-SINDy_alg}) ``streaming" models. 

\paragraph{Fitting method}
\label{par:fitting_method}
We will briefly cover the fitting method used in the experimental section. SINDy and its variants use sparsity promoting techniques to find the coefficient vectors. While LASSO \cite{tibshirani1996regression} is a standard way to promote sparsity, we have opted to use a sequential threshold least squares approach (see Algorithm~\ref{alg:STLTSQ}). In each iteration, if a coefficient is under a user selected threshold $\varepsilon_{\text{coeff}}$, the corresponding column is removed from the feature matrix $\bm{G}$. In this article we will refer to $\varepsilon_{\text{coeff}}$ as the ``coefficient'' threshold. Moreover, we add an $L^2$ regularization term to the fitting problem, i.e. we solve 
\begin{equation}
\bm{c}^* = \operatorname{argmin}_{c}\{\|\bm{G}\bm{c}-\bm{b}\|^2_2  + \lambda\|\bm{c}\|^2_2\}.
\end{equation}
We call $\lambda$ the ``regularization'' parameter. In Algorithm \ref{alg:STLTSQ}, $\bm{c}^*_j$ denotes the $j$-th entry of $\bm{c}^*$, and $\bm{G}_j$ denotes the $j$-th column of $\bm{G}$. We have chosen to use $j$ as the index as it ties to the number of projection functions which determines the length of the coefficient vectors and the number of columns of the feature matrices. 

\begin{algorithm2e}
 \KwInput{threshold $\varepsilon_{\textup{coeff}}$, regularization weight $\lambda$, matrix $\bm{G}$, vector $\bm{b}$}
 \KwResult{Coefficient vector $\bm{c}$}
    Compute $\bm{c}^* = \operatorname{argmin}_{c}\{\|\bm{G}\bm{c}-\bm{b}\|^2_2  + \lambda\|\bm{c}\|^2_2\}$.\\
    \While{There exist $\bm{c}^*_j$ such that $\bm{c}^*_j<\varepsilon_{\textup{coeff}}$}
        {
        Find $j$ such that $\bm{c}^*_j<\varepsilon_{\textup{coeff}}$\\
        Remove $\bm{A}_j$ from $\bm{A}$\\
        Compute $\bm{c}^* = \operatorname{argmin}_{c}\{\|\bm{G}\bm{c}-\bm{b}\|^2_2  + \lambda\|\bm{c}\|^2_2\}$.
        }
    \Return{$\bm{c}^*$}
 \caption{Sequential threshold least squares}
 \label{alg:STLTSQ}
\end{algorithm2e}

\subsection{Lorenz data -- impact of streaming integration}
\label{subsec:Lorenz data}

In this experiment, the Lorenz system 
\begin{equation}
\label{eq:Lorenz_system}
    \dot{\bm{u}}(t) = 
    \begin{bmatrix}
    10 \cdot (u_2-u_1),&
    u_1 \cdot (28-u_3)-u_2,&
    u_1 u_2 - (8/3)\cdot u_3
    \end{bmatrix}^\top
\end{equation}
was used as a source of data, in which 10,001 snapshots over the time interval $[0,10]$ were generated. To demonstrate the effect of the streaming integration on the weak-SINDy process, we calculate a streaming model and a static model from this entire data set. As the state dimension is low, we do not incorporate a POD reduction in this experiment. Let $(\bm{b}_{\text{static}},\bm{G}_{\text{static}})$ and $(\bm{b}_{\text{stream}},\bm{G}_{\text{stream}})$ be the target vectors and feature matrices for the static and streaming models, respectively. For both models, monomials of max degree one were used as a projection basis and the Fourier basis with $T = 10$ and $\tilde{K}=20$. For fitting we used a standard sequential threshold least squares approach (i.e. $\lambda = 0$) with $1\text{e}-1$ as the coefficient threshold $\varepsilon_{\text{coeff}}$ in Algorithm~\ref{alg:STLTSQ}. 

\begin{figure}
\includegraphics[width = \textwidth]{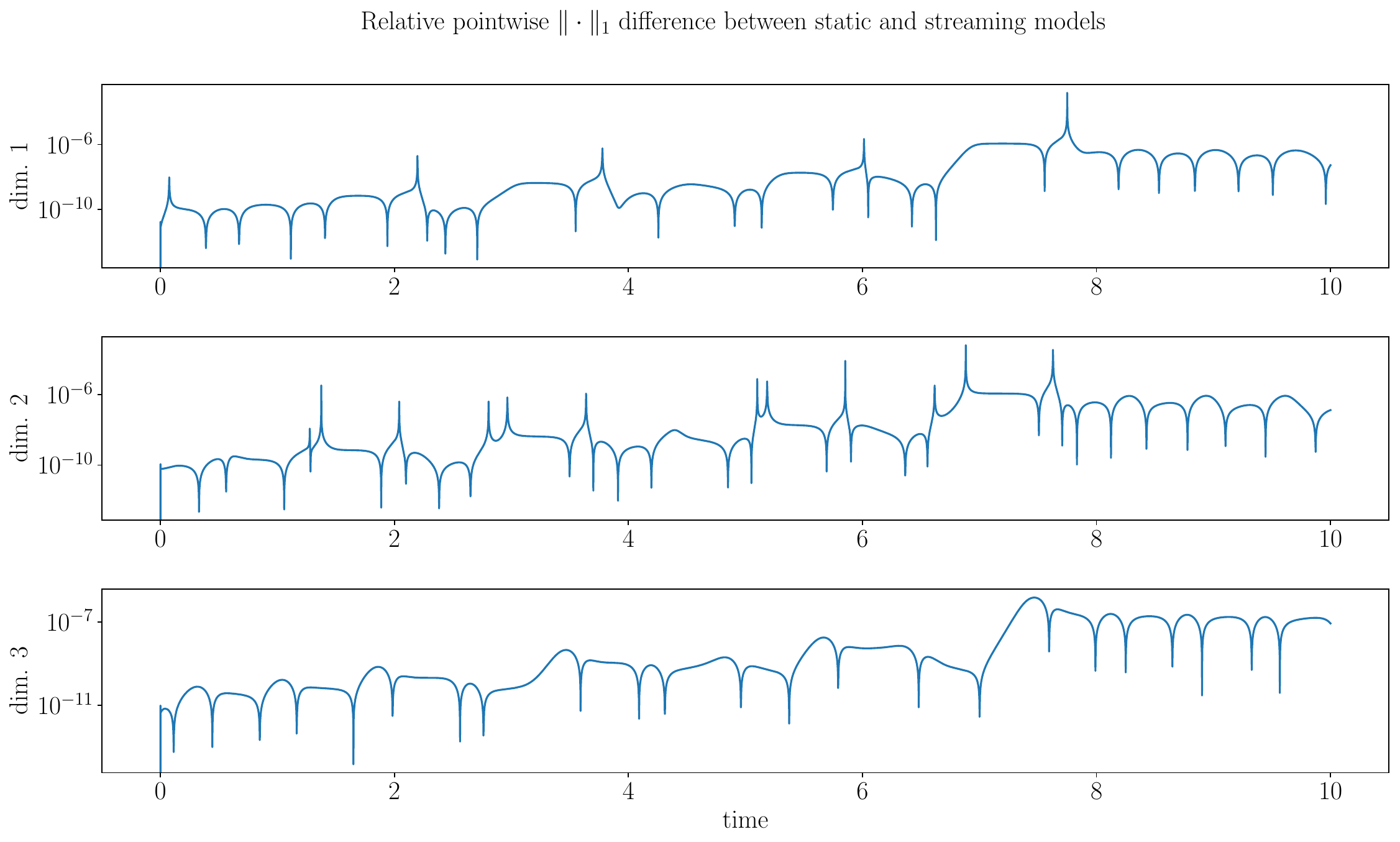}
 \caption{For each state dimension, we have displayed the relative pointwise $\| \cdot \|_1$ difference between trajectories generated by the static model and streaming model. This is defined as $100\cdot\|u_{\text{static}}(t) - u_{\text{streaming}}(t)\|_1/ \|u_{\text{static}}(t)\|_1$, where $u_{\text{static}}$ is the trajectory generated from the static model and $u_{\text{streaming}}$ is the trajectory generated from the streaming model.}
\label{fig:difference_between_static_and_streaming}
\end{figure}

\begin{figure}
\centering
\includegraphics[width = \textwidth]{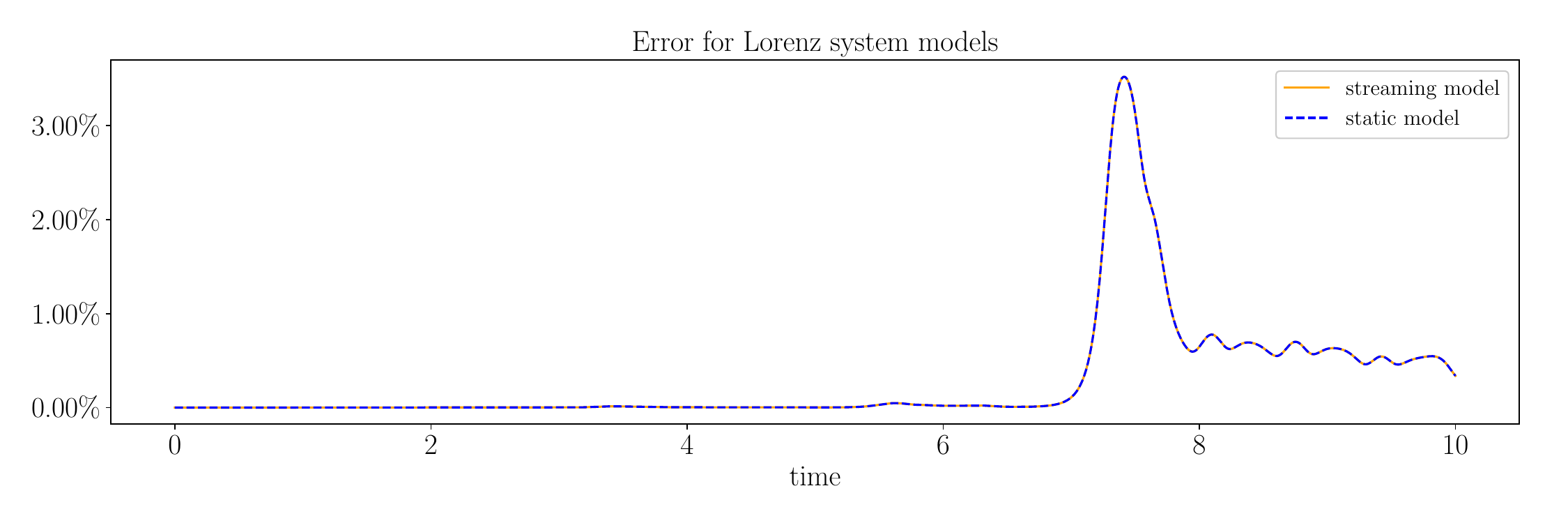}
 \caption{If $\bm{\bm{u}}^*$ is the trajectory from the model and $\bm{\bm{u}}$ is the true trajectory, the $L^2$ percent error is calculated by $E(t)  = 100 \cdot \|\bm{\bm{u}}(t)^* -  \bm{\bm{u}}(t)\|_2/\|\bm{\bm{u}}(t)\|_2$. This figure displays the $L^2$ percent error for the streaming model (solid blue line) and static model (dashed orange line). This demonstrates that although the systems slightly vary, there is no considerable difference in accuracy.}
 \label{fig:percent_error_lorenz}
\end{figure}

\begin{figure}
\centering
\begin{subfigure}{\textwidth}
\includegraphics[width = \textwidth]{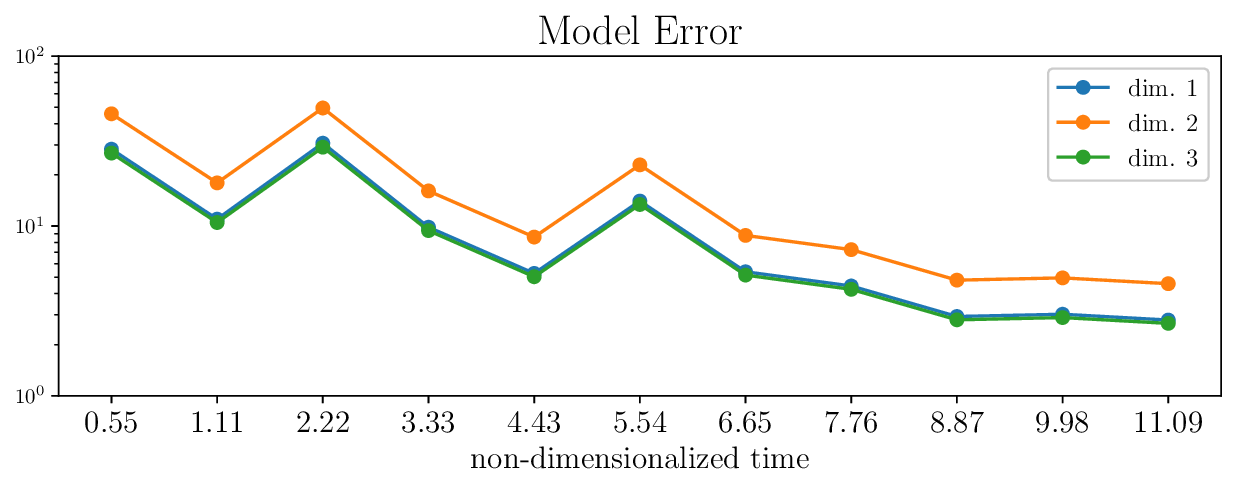}
 \caption{Let $\bm{\bm{u}}^*$ be the trajectory from the model and $\bm{\bm{u}}$ be the true trajectory, the model error in each state dimension is calculated by $E_i  = 100 \cdot \|u^*_i -  u_i\|_\infty/\|u_i\|_\infty$, $i=1,2,3$, where the supremum is taken over the entire time domain. This figure displays the values of $E_i$'s as more snapshots become available over time. In this plot, the $x$-axis is $\frac{t}{t_{\textup{L}}}$ with $t_{\textup{L}}\approx .902$ the Lyapunov time for the Lorenz system, and the $y$-axis is on a logarithmic scale.}%
 \label{fig:error_goes_down_as_samples_go_up_in_time}
 \end{subfigure}
 \begin{subfigure}{\textwidth}
 \includegraphics[width = \textwidth]{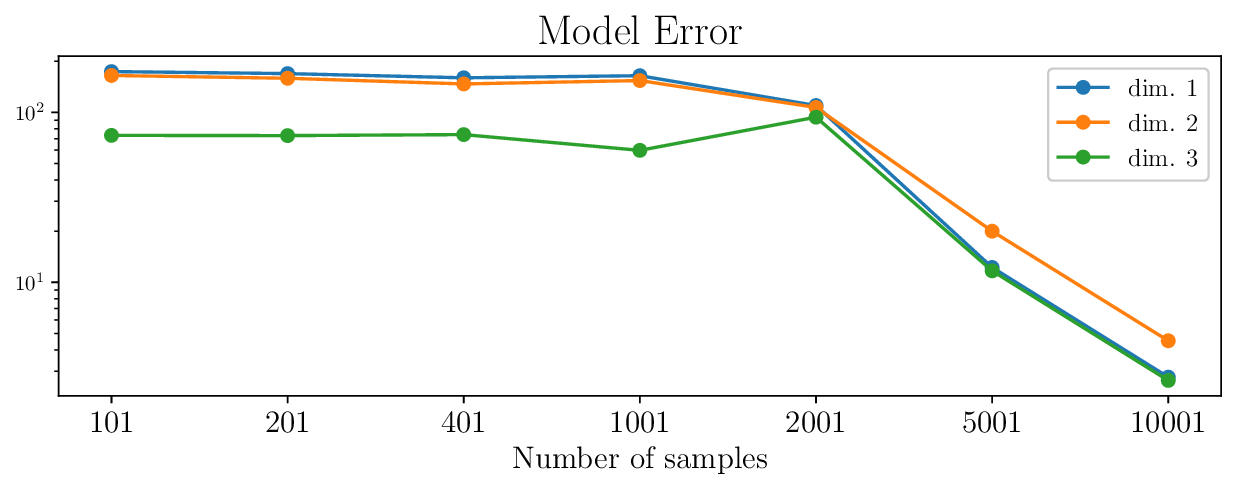}
 \caption{This figure displays the model error $E_i$'s in state dimension $i=1,2,3$, as the density of snapshots increases. In this plot, the $x$-axis is the total number of snapshots taken uniformly over the entire time domain, and the $y$-axis is on a logarithmic scale.}%
 \label{fig:error_goes_down_as_sample_density_goes_up}
 \end{subfigure}
 \caption{Weak-SINDy model error versus the number of available data.}
\label{fig:error_goes_down_as_samples_go_up}
\end{figure}

The streaming integration process outlined in Section \ref{subsec:streaming_integration} results in slightly different integral values from the ones calculated from by quadrature function {\texttt{trapezoid}} in the Python package {\texttt{scipy}}. These slight variations on the feature matrix and target vectors can accumulate over time. Figure \ref{fig:difference_between_static_and_streaming} displays a relative pointwise $L^1$ difference between trajectories generated from the static and streaming models. The results show that the data reconstructed from the streaming model are slightly different from the ones reconstructed from a static model, due to the different integration methods. However, Figure \ref{fig:percent_error_lorenz} shows that the data reconstructed from both the streaming and the static models have nearly identical pointwise $L^2$ errors when compared to the original data from the Lorenz system (Eq.~\eqref{eq:Lorenz_system}). 
Together, the results in Figures \ref{fig:difference_between_static_and_streaming} and \ref{fig:percent_error_lorenz} confirm that the streaming integration procedure has minimal effect on the overall accuracy of the reconstructed data. Finally, we computed the sup-norm percent error for each state dimension of data reconstructed from a weak-SINDy model as the number of snapshots increases and report the results in Figure~\ref{fig:error_goes_down_as_samples_go_up}.  
In Figure~\ref{fig:error_goes_down_as_samples_go_up_in_time}, we report the model error with respect to the number of snapshots collected over time. Here we plot in terms of the non-dimensionalized time $\frac{t}{t_{\textup{L}}}$, where $t_{\textup{L}} \approx 0.902$ is the Lyapunov time for the Lorenz system computed by using python package \texttt{lyapynov} \cite{Lyapunov}. We observe that the error decreases as data from later time becomes available. However, the benefit diminishes when sufficient data are collected. Figure~\ref{fig:error_goes_down_as_sample_density_goes_up} plots the modeling error as the total number of snapshots over the entire time interval increases, i.e., time-step size decreases. Here the $x$-axis depicts the total number of snapshots. In general, the results indicate that the data reconstruction error decreases as the number of snapshots increases. Hence, the compression efficiency of the streaming weak-SINDy method improves as the number of snapshots increases and at no additional memory cost, since the memory footprint of the method is independent of the number of snapshots. Table \ref{table:summary_storage_requirements_lorenz} shows the storage requirements for the streaming weak-SINDy method applied to the compression of the Lorenz data set. In this relatively simple example, the compression is extremely efficient. Online, the method only stores the feature matrix and target vectors, which combined is 1.5\% of the total size of the entire data set. Offline, we store the coefficients, which is .08\% the size of the data set. %
\begin{table}[htp]
\centering
\renewcommand{\arraystretch}{1}
\begin{tabular}{r|c|c}
&no compression  & streaming weak-SINDy\\
\hline
Online storage & $3\times 10001$ & \makecell{ $(41\times 8) + (41\times 3)$ -- feature matrix and\\ \hspace{2.3cm} target vector} \\
\hline
Offline storage & $3\times 10001$ & $(8\times 3)$ -- coefficients
\end{tabular}
\caption{Storage requirements for streaming weak-SINDy compression applied to the Lorenz data set.}
\label{table:summary_storage_requirements_lorenz}
\end{table}%

\subsubsection{Effects of sparsity enforcement}
\label{subsubsec:Effects_of_sparsity}
 In this subsection, we investigate the effects of sparsity enforcement in the weak-SINDy surrogate models. While sparsity can reduce the sizes of the coefficient vectors and the associated bases in the weak-SINDy surrogates, the saving from enforcing sparsity is relatively small when compared to the compression of time-dependent data into a (potentially non-sparse) regression-based surrogate model.
 We construct a weak-SINDy surrogate model without enforcing sparsity in the regression process by setting the sparsity threshold $\varepsilon_{\textup{coeff}}=\infty$ (which leads to a standard $L^2$ regularization) and compare the model error of this dense model to the sparse model considered in Section~\ref{subsec:Lorenz data}. This numerical experiment results in 
\begin{equation}
\left[\frac{e_{1,\textup{d}}}{e_{1,\textup{s}}},\, \frac{e_{2,\textup{d}}}{e_{2,\textup{s}}},\, \frac{e_{3,\textup{d}}}{e_{3,\textup{s}}} \right] = [27.87,\,   4.88,\, 229.87],
\end{equation}
where $e_{i,\textup{d}}$ and $e_{i,\textup{s}}$ denote the $\ell_2$ difference between the Lorenz system parameters in Eq.~\eqref{eq:Lorenz_system} and the dense and sparse surrogate coefficients for dimension $i$, respectively. 
The result shows that sparsity enforcement leads to more accurate coefficient recovery, which agrees with observations reported for SINDy in \cite{brunton2016discovering} and is the primary reason that sparsity enforcement is included in the proposed approach. 

\subsection{Fluid flow}
\label{subsec:Fluid_Flow}
For a numerical demonstration, simulated fluid flow data is created using a Lattice Boltzmann method enacted through Python code found at \cite{Lattice_boltzmann}. In this implementation, the lattice spacing is set to $\Delta x = 1$ and the time step is taken to be $\Delta t = 1$. The simulation was ran for 100,000 time steps and a portion of 10,000 time steps between snapshots $32,000$ and $42,000$ was then selected to be considered as the data set in this test. This choice is made to guarantee that the data considered is not subject to rapid change. The case when the data experiences rapid changes is discussed later in Section~\ref{subsubsec:Deficiency}. In this experiment, Algorithm \ref{alg:weak-PSINDy} was tested on the selected 10,000 snapshots of fluid flow data. For the POD update function (Function \ref{func:streaming_POD}), the spectral threshold $\varepsilon_{\text{spec}}$ was set to $1\text{e}-1$ and the residual threshold $\varepsilon_{\text{res}}$ was set to $10\%$. From the initial 550 snapshots, the streaming POD process generated an initial $14$ spatial modes, i.e., $p_0=550$ in Algorithm~\ref{alg:weak-PSINDy}. The streaming POD process then added two additional POD modes at $n=585$ and $n= 636$. For the streaming weak-SINDy process in Algorithm \ref{alg:weak-PSINDy}, monomials of total degree two were chosen for projection functions and a Fourier basis with $\tilde{K} = 99$ ($K=2\tilde{K}+1=199$) and $T = 10,000$ for the test functions. To solve the regression problem, Function \ref{func:Build_and_Solve} was applied, where each coefficient vector was solved for using a sequential threshold least squares approach with an $L^2$ regularization term (see Algorithm \ref{alg:STLTSQ}). For the fitting method parameters see Table \ref{table:fitting_parameters}.

\begin{table}[htp]
\centering
\renewcommand{\arraystretch}{1} 
\begin{tabular}{l|c|c}
\ &regularization weight & coefficient threshold \\
\hline
coefficient vectors 1 -- 14 & $\sim 1.600\text{e}-06$ &0.0003 \\
\hline
coefficient vector 15 &$\sim 5.791\text{e}-08$& 0.0004\\
\hline
coefficient vector 16 &$\sim 2.763\text{e}-09$& 0.0004\\
\end{tabular}
\caption{Fitting parameter values used in the sequential threshold least squares method for determining coefficient vectors associated to the first 14, the 15th, and the 16th POD modes.}
\label{table:fitting_parameters}
\end{table}

\begin{figure}
\centering
\includegraphics[width = .5\textwidth]{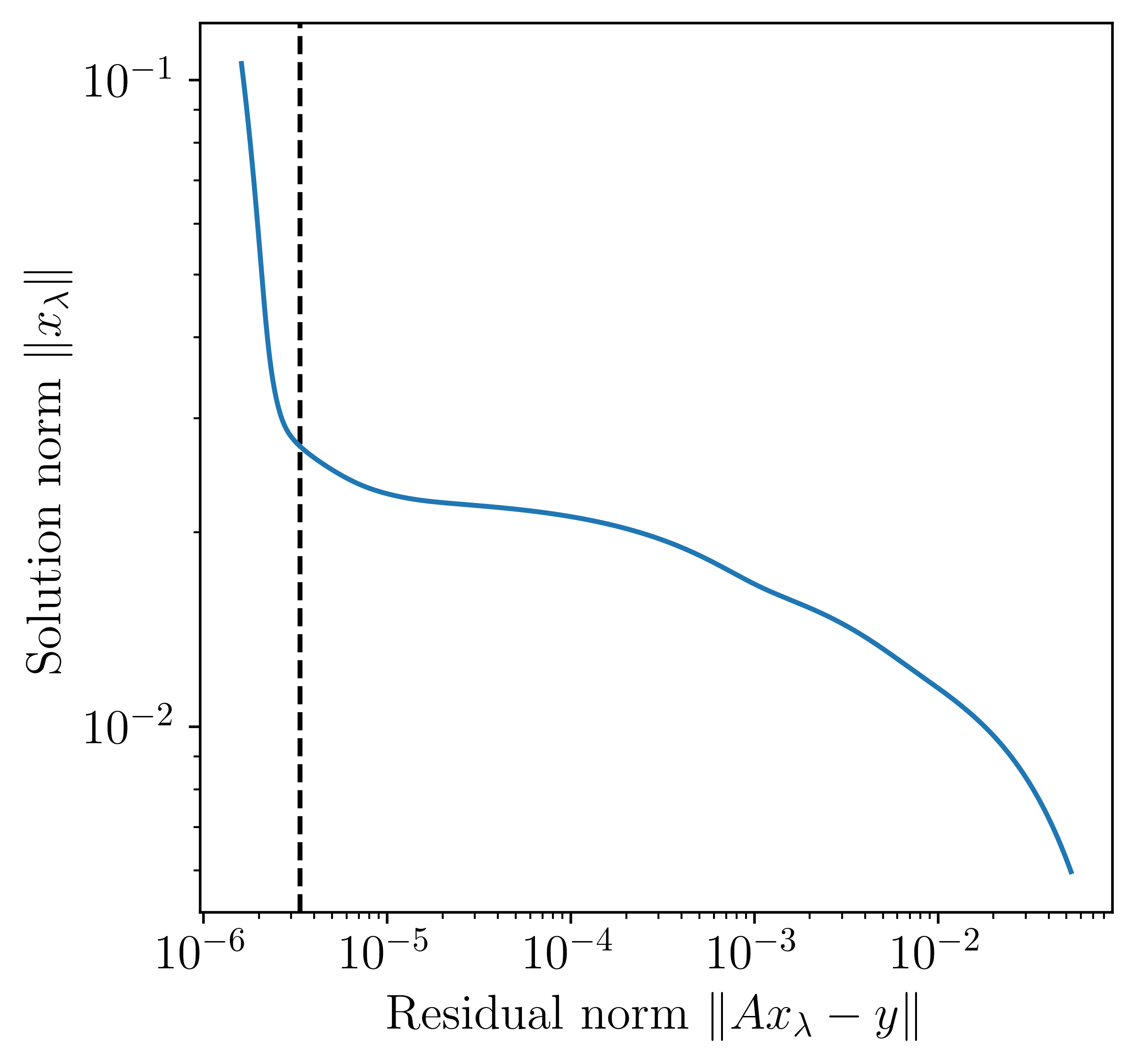}
\caption{An example L-curve for choosing the regularization weights $\lambda$ in Section~\ref{subsec:Fluid_Flow}. The dashed vertical line represents the choice of $\lambda$.}
\label{fig:l-curve}
\end{figure}

\begin{figure}
\begin{subfigure}{\textwidth}
\centering
\includegraphics[width = \textwidth]{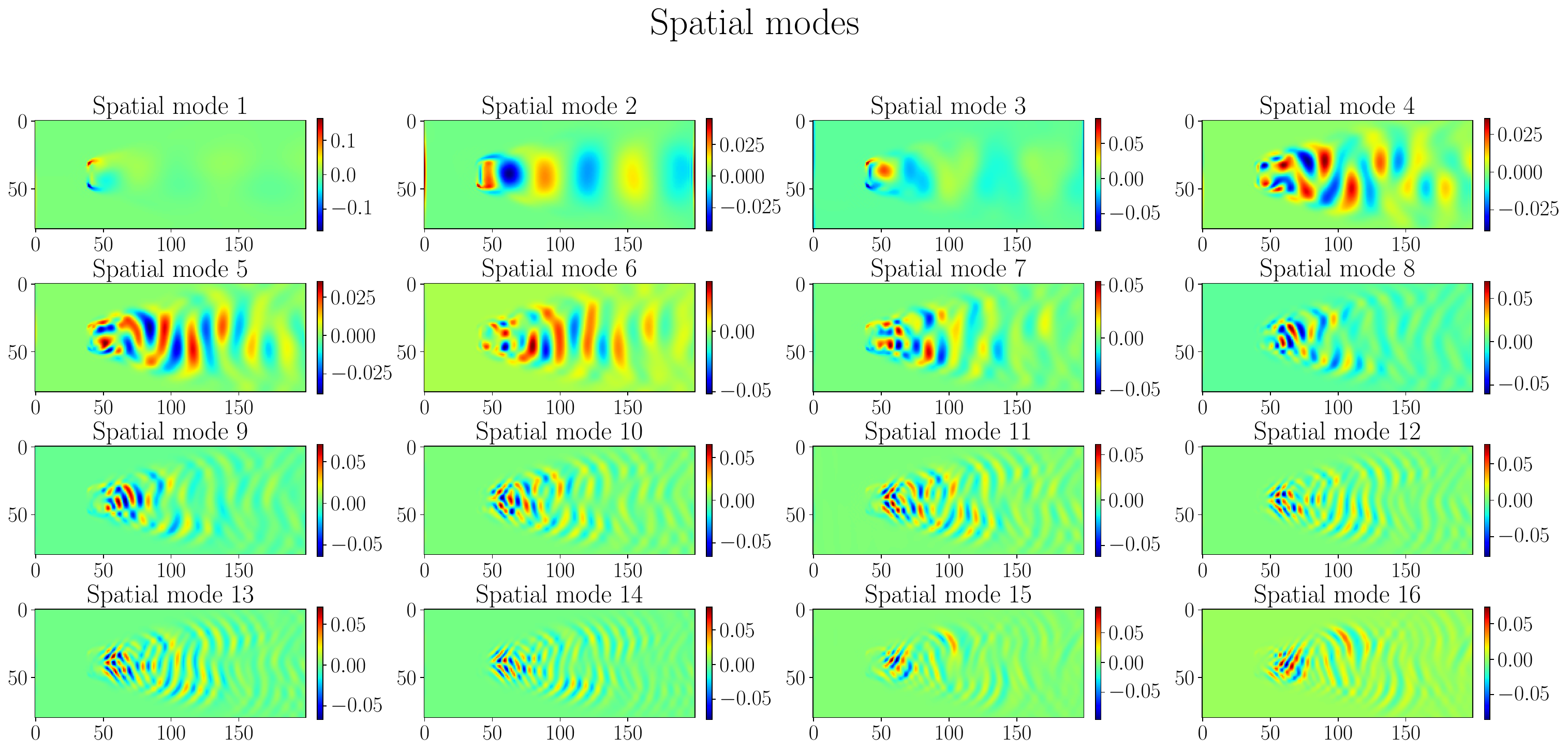}
 \caption{Spatial modes created from the streaming POD process. The two modes on the bottom right are the added modes.}
 \label{fig:spatial_modes}
\end{subfigure}

\begin{subfigure}{\textwidth}
\centering
\includegraphics[width = \textwidth]{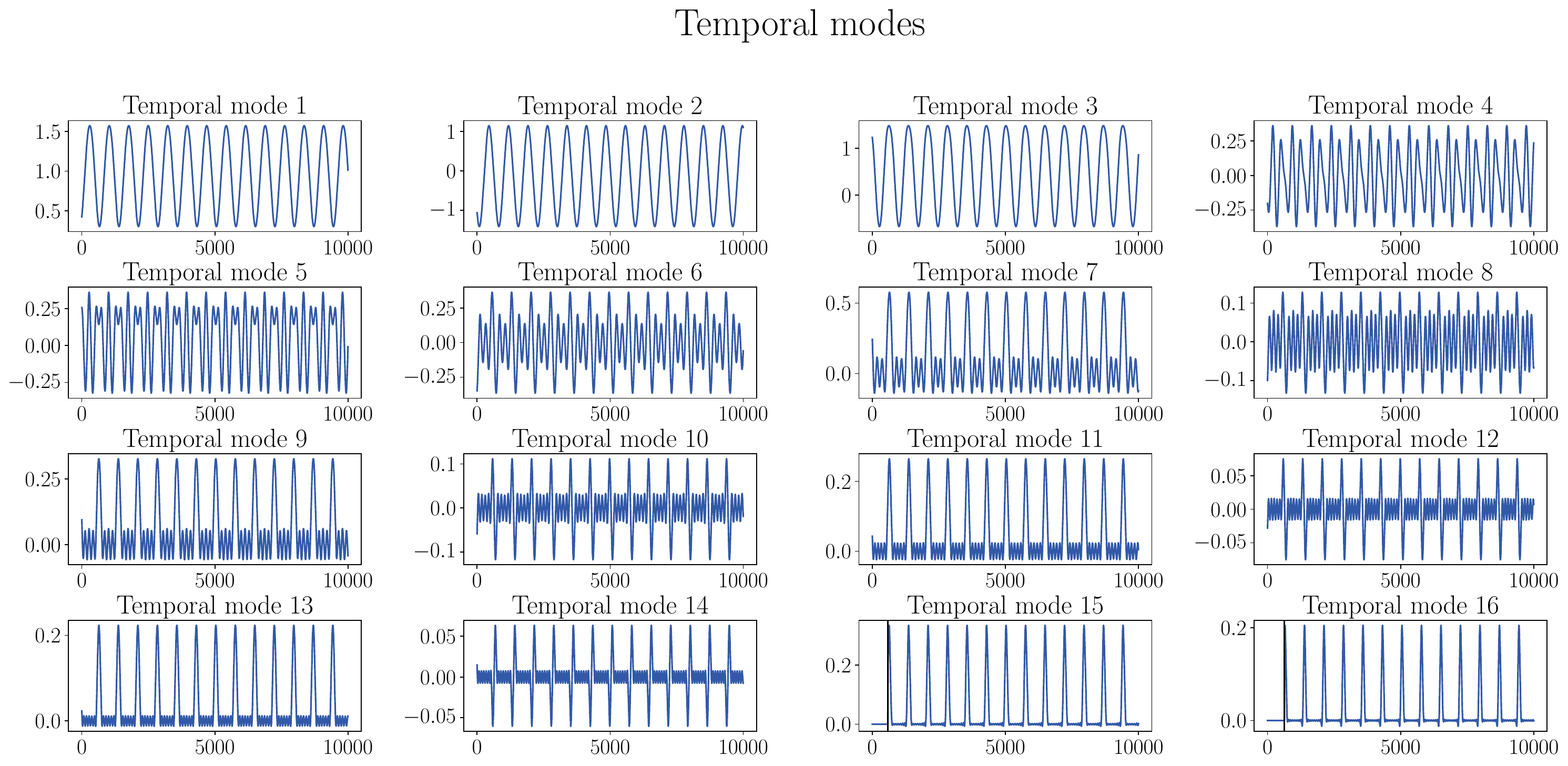}
 \caption{The temporal modes computed using the streaming integration process. The last two temporal modes on the bottom right start at a later point than the other, this is indicated by a black vertical line at the start point.}
 \label{fig:temporal_modes}
 \end{subfigure}%
 \caption{The spatial and temporal streaming POD modes for the fluid flow data.}
 \label{fig:streaming_POD_modes}
\end{figure}

\begin{figure}
\centering
\includegraphics[width = \textwidth]{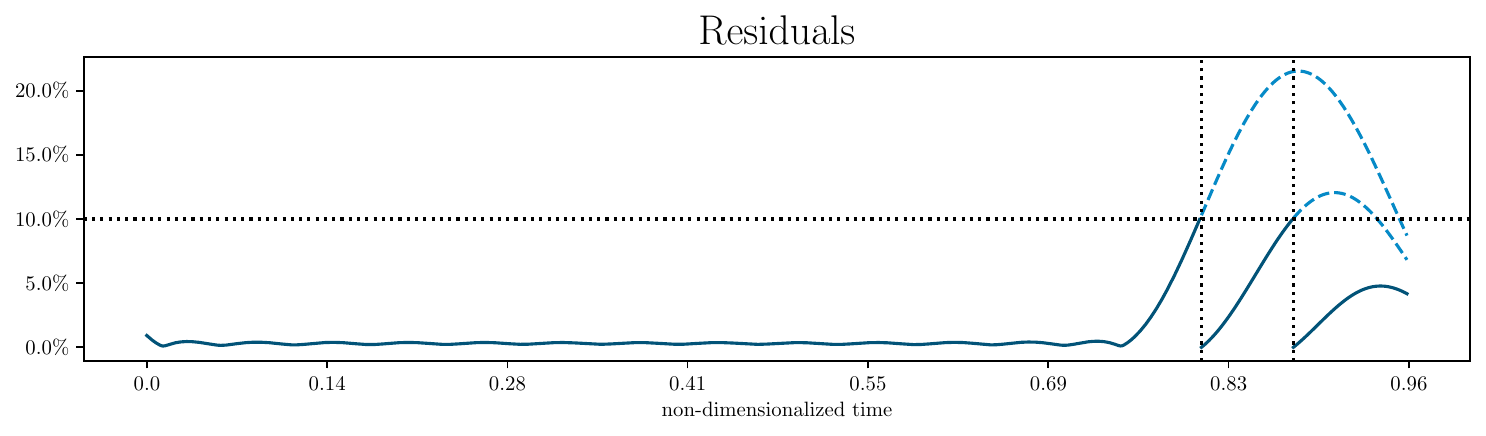}
\caption{The solid blue line displays the streaming POD residual (Equation~\eqref{eq:residual}) for each snapshot $\bm{u}(\hat{\bm{x}}, t_n)$ over the first 700 snapshots. New spatial modes were added at $n = 585$ and $n = 636$ as indicated by the dotted vertical lines. The threshold was set to 10\% indicated by the dotted horizontal line. The light blue dashed line shows the value of the residual if the new modes were not added.The $x$-axis displays the non-dimensionalized time $\frac{t}{t_{\textup{V}}}$ with $t_{\textup{V}}\approx727$ the vortex shedding period for this problem.}
\label{fig:residuals}
\end{figure}

\begin{figure}
\centering
\includegraphics[width = \textwidth]{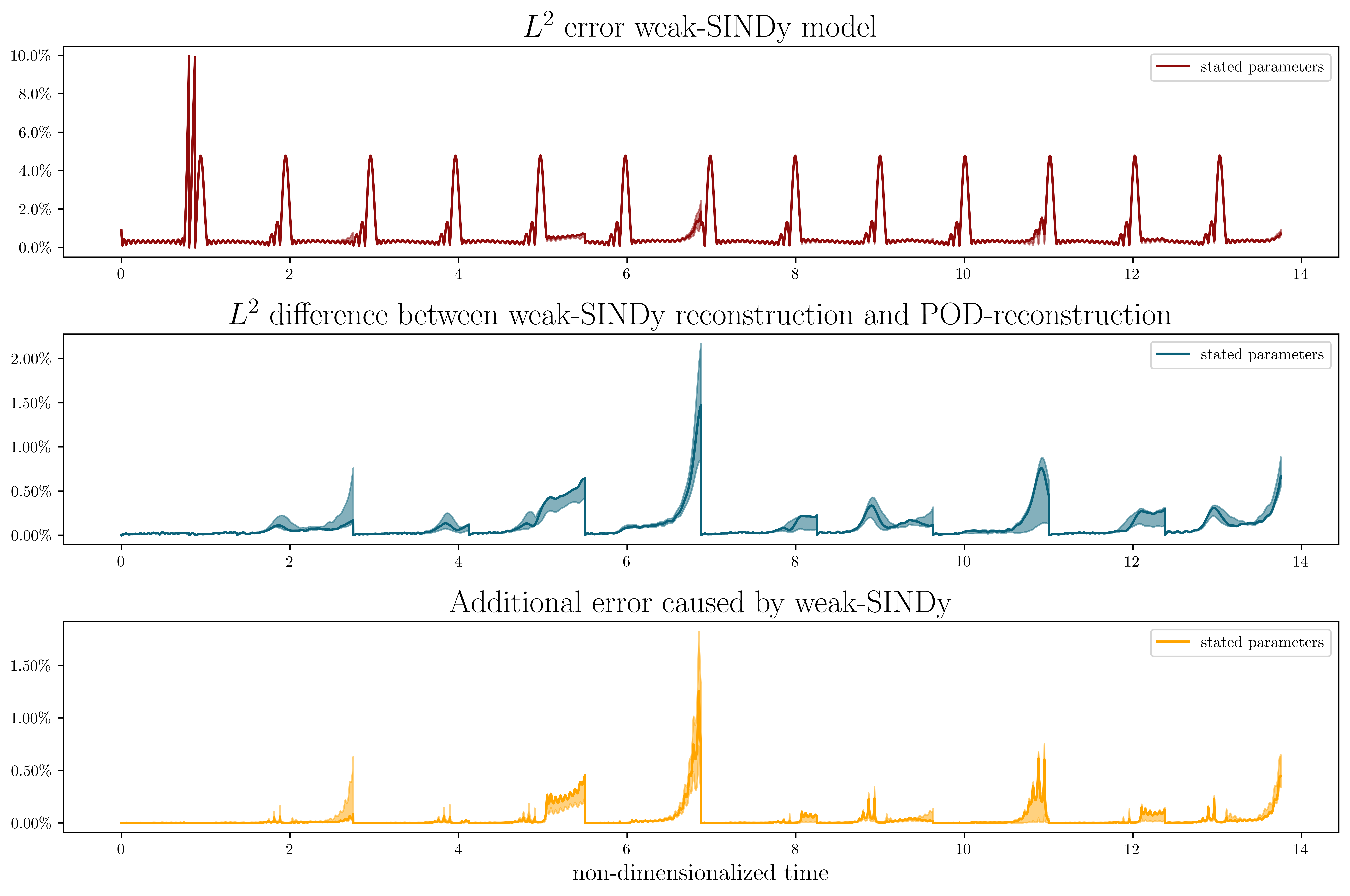}
\caption{Top plot shows the data reconstruction error from the weak-SINDy model $u(\bm{x},t)$. The error is defined as $E(t) = 100 \cdot \|u(\bm{x},t) - \tilde{u}(\bm{x},t)\|_\Omega / \|u(\bm{x}, t) \|_\Omega$. The threshold for the streaming POD process was set to $10 \%$ and new POD modes where added at time steps $n = 585$ and $636$. 
We have also calculated the percent difference between the streaming POD reconstruction and the weak-SINDy reconstruction (middle plot) and the additional error caused by weak-SINDy (bottom plot). See Eqs.~\eqref{eq:percent_diff_pod_wkSINDy} and \eqref{eq:error_caused_by_wkSINDy}. In each plot, the shaded region shows error variations resulting from a $10\%$ perturbation on the regularization weight $\lambda$. The $x$-axes in this figure display the non-dimensionalized time $\frac{t}{t_{\textup{V}}}$ with $t_{\textup{V}}\approx727$ the vortex shedding period for this problem.}
\label{fig:Error_reconstruction}
\end{figure}

\begin{figure}
\centering
\includegraphics[width = \textwidth]{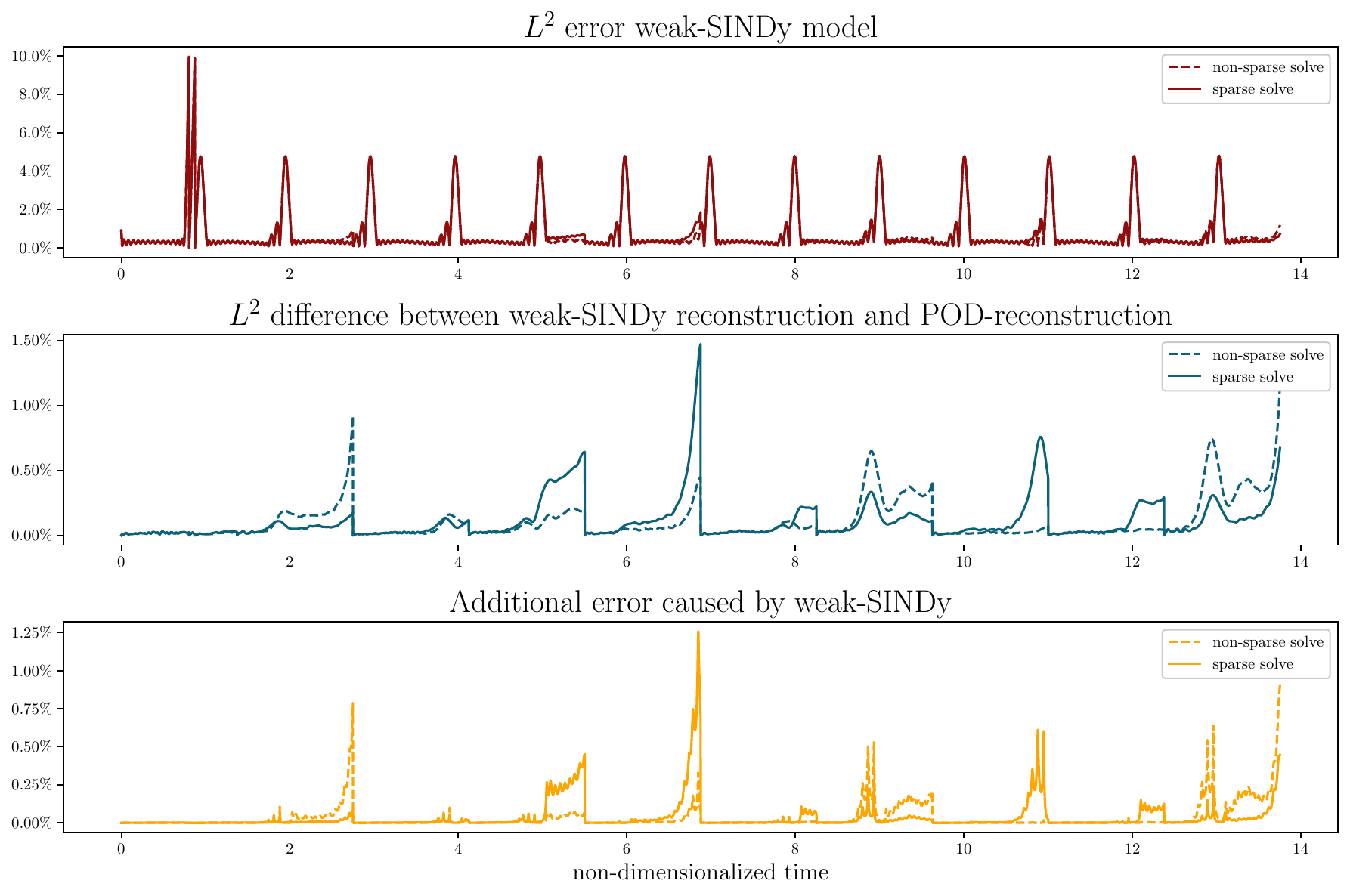}
\caption{The comparison between model errors for surrogate models constructed with and without sparsity enforcement in the regression process. The reported model errors are calculated following the same approach as the one reported in Figure~\ref{fig:Error_reconstruction}, with the solid lines and dashed lines denoting the errors from sparse and dense surrogate models, respectively. 
Here, the $x$-axes display the non-dimensionalized time $\frac{t}{t_{\textup{V}}}$ with $t_{\textup{V}}\approx727$ the vortex shedding period for this problem. }
\label{fig:Error_reconstruction_without_sparsity}
\end{figure}

The regularization weights $\lambda$ were chosen via the L-curve technique \cite{hansen1999curve}, and the coefficient thresholds were chosen in a similar fashion. An example L-curve is presented in Figure \ref{fig:l-curve}. Specifically, after the regularization weight is chosen, the coefficient thresholds are chosen to be large enough to promote some sparsity of the coefficients while maintaining similar least squares residuals to the non-thresholded case. Finally, in addition to the data recorded in Algorithm \ref{alg:weak-PSINDy}, we recorded the snapshots of the temporal modes $\hat{\nu} = [\nu_1(t_n),\ldots, \nu_{16}(t_n)]^\top$ every 1000 time steps to serve as restart points for the model resulting from weak-SINDy in order to counter the error accumulation in time evolution and improve the data reconstruction accuracy.%
Figure \ref{fig:streaming_POD_modes} displays the spatial modes created by Function \ref{func:streaming_POD} and the resulting temporal modes calculated in a streaming process by Eq.~\eqref{eq:temporal-modes}. On the last two temporal modes, we have indicated via a vertical black line the time step at which the corresponding spatial mode was added. Figure \ref{fig:residuals} displays the values of the residual (see Eq.~\eqref{eq:residual}) over the first 700 snapshots. Again, we have indicated where the additional spatial POD modes were added with dotted vertical lines. The residual threshold of $10\%$ is indicated with a horizontal dotted line. Moreover, the value of the residual if the additional spatial mode was not added is indicated for reference. For the data $\tilde{u}(\bm{x},t)$ reconstructed using models from Algorithm \ref{alg:weak-PSINDy}, some additional error is incurred in the regression process in Function \ref{func:Build_and_Solve} for the temporal modes. Figure~\ref{fig:Error_reconstruction} illustrates the model errors over the non-dimensionalized time $\frac{t}{t_{\textup{V}}}$, where $t_{\textup{V}}\approx727$ is the vortex shedding period for this problem. 
The top plot in Figure~\ref{fig:Error_reconstruction} shows the overall percent error $E(t) = 100 \cdot \|\tilde{u}(\bm{x},t) - u(\bm{x},t)\|_\Omega / \| u(\bm{x},t)\|_\Omega$ in red.
The middle plot reports the difference between the reconstructed data $\tilde{u}(\bm{x},t)$ and the POD expansion $u_{\text{pod}}(\bm{x},t)$ calculated by 
\begin{equation}
\label{eq:percent_diff_pod_wkSINDy}
D(t) = 100 \cdot ((\|\tilde{u}(\bm{x},t) - u_{\text{pod}}(\bm{x},t)\|_\Omega )/ \|u_{\text{pod}}(\bm{x},t)\|_\Omega),
\end{equation}
 displayed in blue. Finally, the bottom plot in Figure \ref{fig:Error_reconstruction} shows the percent error caused by fitting the surrogate to the POD approximation calculated by 
\begin{equation}
\label{eq:error_caused_by_wkSINDy}
E_w(t) = 100 \cdot ( {(\|\tilde{u}(\bm{x},t) - u(\bm{x},t)\|_\Omega - \|u(\bm{x},t) - u_{\text{pod}}(\bm{x},t)\|_\Omega)/\| u(\bm{x},t)\|_\Omega}).
\end{equation}
Moreover, the value of the regularization weight $\lambda$ has been varied by $\pm10\%$ from the ones reported in Table~\ref{table:fitting_parameters} and the resulting variations in the errors are shown as the shaded regions in Figure~\ref{fig:Error_reconstruction}. 
We can see that a majority of the surrogate model error is caused by the accuracy of the POD approximation and that the weak-SINDy model is robust under perturbation of the regularization weights. In addition, we investigate the effect of sparsity in the fluid flow surrogate model following the approach in Section~\ref{subsubsec:Effects_of_sparsity} for the Lorenz system and report the results in Figure~\ref{fig:Error_reconstruction_without_sparsity}. 
These results indicate that the effect of sparsity-enforcing regression is insignificant compared to the POD approximation error. Finally, we observe that after the number of POD modes stops increasing, the reconstruction error of the weak-SINDy model stays roughly constant as more snapshots become available over time. This is due to the fact that, in this fluid flow example, the weak-SINDy reconstruction error is mostly due to the POD approximation error.%

\subsubsection{Analysis of the compression of fluid flow data}
\label{subsubsec:analysis_fluidflow}
The overall size of the fluid flow data set is given in Table \ref{table:size_data_set}, which indicates that, storing the entire fluid flow data set requires recording $160,000,000$ entries. When the streaming POD process is applied to the data set, this results in $16$ spatial modes and $16$ temporal modes. However, the last two modes of the spatial modes and temporal modes were created at time steps $n=585$ and $n=636$. Given the size of POD modes specified in Table~\ref{table:size_of_streamed_POD_modes}, storing the spatial POD modes requires $256,000$ entries and storing the temporal modes requires $158,779$ entries. At the start of the streaming POD process, an initial $550$ snapshots was temporarily recorded for the generation of the initial POD modes, which requires temporarily storing $8,800,000$ non-zero entries based on Table \ref{table:size_data_set}.

For Algorithm \ref{alg:weak-PSINDy}, we temporarily store the target vectors and feature matrix before solving the regression problems. Tables \ref{table:form_feature_matrix_data_vec} and \ref{table:blockmatrices_entry_sizes} show this requires saving $90,346$ entries accounting for the non-zero matrices in the block form of $\bm{G}$, and the target vectors. Solving the regression problems given by the feature matrix and target vectors results in $16$ coefficient vectors, each of length $153$ (one coefficient per projection function per temporal mode). Thus, as the data can be reconstructed by evolving the model from the coefficient vectors, it is only necessary to store the $2,448$ entries of the coefficient vectors. 

The above discussion is summarized in Table \ref{table:summary_storage_requirements}. In the example of the fluid flow data set, Table \ref{table:summary_storage_requirements} shows that Algorithm \ref{alg:weak-PSINDy} is more memory efficient than the streaming POD process alone in terms of both online and offline storage requirements.
We note that further compression may be attained by leveraging structural properties of the data, such as symmetry of POD modes for the fluid data considered here, which could also lead to better reconstruction accuracy.
\begin{table}
\centering
\renewcommand{\arraystretch}{1}
\begin{tabular}{ r | c | c}
\ & {\bf number of snapshots} & {\bf size} \\
\hline
{\bf fluid flow data set} & $10,000$ & $80\times 200$\\
\end{tabular}
\caption{Size of the fluid flow data set.}
\label{table:size_data_set}
\end{table}
\begin{table}
\centering
\renewcommand{\arraystretch}{1} 
\scalebox{.85}{\begin{tabular}{c | c | c | c}
\thead{feature matrix} & \thead{target vector\\ temporal modes 1-14} & \thead{target vector\\ temporal mode 15} & \thead{target vector\\ temporal mode 16}\\
\hline
$\bm{G} = \begin{bmatrix}
 \bm{G}_1 & \bm{0} & \bm{0}\\
 \bm{G}_2 & \bm{G}_4 & \bm{0}\\
 \bm{G}_3 & \bm{G}_5 & \bm{G}_6\\
\end{bmatrix}$

& 

$\bm{b} = \begin{bmatrix}\bm{b}_1^\top & \bm{b}_2^\top & \bm{b}^\top_3 \end{bmatrix}^\top$

& 

$\bm{b} = \begin{bmatrix} \bm{b}^\top_1 & \bm{b}^\top_2 \end{bmatrix}^\top$

&

$\bm{b} = \bm{b}_1$\\

\end{tabular}}
\caption{A table of the block forms of the ``Build" portion of \texttt{BuildAndSolve}. Here, the indexing does \emph{not} correspond to the indexing presented in the pseudo code but instead count the blocks.}
\label{table:form_feature_matrix_data_vec}
\end{table}%
\begin{table}[htp]
\begin{minipage}{.5\linewidth}
\centering
\begin{tabular}{r|c}
\multicolumn{1}{c|}{{\bf matrix}} & {\bf size} \\
\hline
$\bm{G}_1$, $\bm{G}_2$, $\bm{G}_3$ & $199 \times 120$\\
\hline
$\bm{G}_4$, $\bm{G}_5$  &  $199 \times 16$\\
\hline
$\bm{G}_6$ & $199 \times 17$\\
\hline
$\bm{b}_1$, $\bm{b}_2$, $\bm{b}_3$ & $199 \times 1$\\
\end{tabular}
\caption{Size of the matrices in Table \ref{table:form_feature_matrix_data_vec}.}
\label{table:blockmatrices_entry_sizes}
\end{minipage}%
\begin{minipage}{.5\linewidth}
\centering
\begin{tabular}{r|c}
\multicolumn{1}{c|}{{\bf modes}} & {\bf size} \\
\hline
spatial modes 1-16 & $80\times 200$ each \\
\hline
temporal modes 1-14  &  $1\times 10,000$ each \\
\hline
temporal modes 15 & $1\times 9,415$\\
\hline
temporal modes 16 & $1\times 9,364$\\
\end{tabular}
\caption{POD mode sizes from the streaming POD algorithm.}
\label{table:size_of_streamed_POD_modes}
\end{minipage}
\end{table}%
\begin{center}
\begin{table}[htp]
\centering
\renewcommand{\arraystretch}{1}
\scalebox{.9}{\begin{tabular}{r|c|c|c}
&no compression & streaming POD & \makecell[l]{streaming\\ weak-SINDy and POD}\\
\hline
\makecell[c]{Online\\ storage} & 160,000,000 & \makecell[l]{8,800,000 -- snapshots (temp.)\\ 256,000 -- spatial modes\\ 158,779 -- temporal modes} & \makecell[l]{8,800,000 -- snapshots (temp.)\\ 256,000 -- spatial modes\\ 90,346 -- $\bm{G}$ and $\bm{b}$} \\
\hline
\makecell[c]{Offline\\ storage} & 160,000,000 & \makecell[l]{256,000 -- spatial modes\\ 158,779 -- temporal modes} & \makecell[l]{256,000 -- spatial modes\\ 2624 -- $\bm{c}$ and initial data}
\end{tabular}}
\caption{Summary of the storage requirements for the standalone streaming POD and the one combined with weak-SINDy (Algorithm~\ref{alg:weak-PSINDy}). Here the values represent the maximum number of non-zero entries which must be stored. The denotation of ``temp'' refers to the fact that the snapshots used to initially generate the POD modes can be discarded afterwards.}
\label{table:summary_storage_requirements}
\end{table}
\end{center}

\subsubsection{Deficiency with the current algorithm}
\label{subsubsec:Deficiency}
\begin{figure}
\centering
\includegraphics[width = \textwidth]{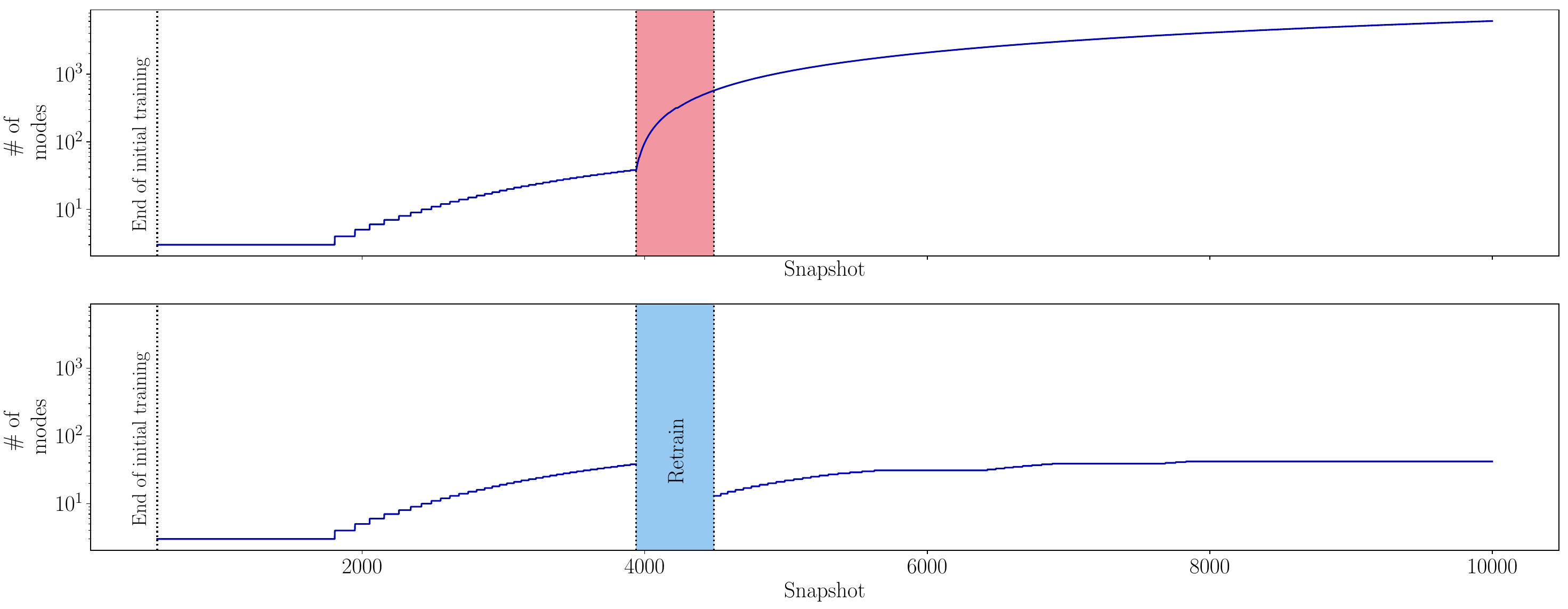}
\caption{The top plot shows the number of spatial POD modes generated at each snapshot using the same parameters for Function \ref{func:streaming_POD} as in Section \ref{subsec:Fluid_Flow} on a different portion of the fluid flow data between snapshots $22,000$ and $32,000$. The highlighted red shaded region shows a period of substantial growth in the number of POD modes. The bottom plot of Figure \ref{fig:POD_explosion} shows the number of POD modes at each snapshot, where $550$ snapshots from the start of the blue shaded region was used to reinitialize in the streaming POD process and the residual threshold $\varepsilon_{\text{res}}$ in Function \ref{func:streaming_POD} was relaxed from 10\% to $15\%$.\vspace{-.5em}}
\label{fig:POD_explosion}
\end{figure}

In this experiment, we examine the streaming POD process on another portion of the $100,000$ time steps between snapshots $22,000$ and $32,000$. This corresponds to a portion of the fluid flow simulation with rapid change. The top plot of Figure \ref{fig:POD_explosion} shows the number of spatial POD modes generated at each time step using the same parameters for Function \ref{func:streaming_POD} as in Section \ref{subsec:Fluid_Flow}. The highlighted red shaded region shows a period of substantial growth in the number of POD modes. 
Here, due to the rapid changes of the data, the initial POD modes from the first $550$ snapshots does not form a good basis for later snapshots, forcing the streaming POD method to increase the number of spatial modes drastically.
As the number of POD modes affects the overall memory footprint of the combined algorithms, the results in the top plot of Figure \ref{fig:POD_explosion} indicate that the efficiency of the streaming weak-SINDy algorithm would be significantly affected by the growing number of POD models.
On the other hand, the bottom plot of Figure \ref{fig:POD_explosion} shows the number of POD modes at each time step when the first $550$ snapshots from the start of the blue shaded region was used to reinitialize the streaming POD process and the residual threshold was relaxed from 10\% to $15\%$. The result here suggests that the issue of growing POD modes for rapidly changing data could be curtailed by a retraining/reinitialization of the streaming POD process. However, this is potentially expensive and warrants future research into either mitigating the expense or identifying systems in which this compression procedure is appropriate.

\section{Conclusion}
In this paper, a streaming weak-SINDy algorithm was proposed. This algorithm extracts the feature matrix and target vectors from streaming data in an online fashion via a streaming integration technique. In an offline process, a regression problem based on the feature matrix and target vectors is then solved to recover a model that governs the evolution of the data. For high-dimensional data sets, a streaming POD process is interjected to act as a dimension reduction technique, which significantly reduces the data dimension of interest while preserving the reconstruction accuracy by adaptively adding new modes. A combination of the streaming POD process and the streaming weak-SINDy algorithm (Algorithm \ref{alg:weak-PSINDy}) then offers further compression on top of the streaming POD dimension reduction by compressing the temporal mode data. The advantages of this combined algorithm are confirmed in two proof-of-concept examples in terms of both online and offline storage costs. In addition, the sizes of feature matrices and target vectors constructed in the streaming weak-SINDy algorithm are independent to the number of streaming data. Therefore, the streaming weak-SINDy algorithm could leverage any additional data to improve the decompression/reconstruction accuracy with minimal effect to the memory cost.
As shown in the discussion, the efficiency of the algorithm depends on the regularity of the streaming data in a number of ways. In particular, when the data rapidly changes, the number of POD modes needed to well-represent the data may outweigh the benefit of the proposed algorithm. The improvement of this process is a potential direction for future research.

\section{Funding}
This work was supported by the Office of Advanced Scientific Computing Research and performed at the Oak Ridge National Laboratory, which is managed by UT-Battelle, LLC for the US Department of Energy under Contract No. DE-AC05-00OR22725. This work was completed during Benjamin Russo's postdoctoral appointment at Oak Ridge National Laboratory.
\FloatBarrier

\bibliographystyle{siamplain}
\bibliography{references.bib}

\begin{thebibliography}{10}

\bibitem{archibald2022dictionary}
{\sc R.~Archibald and H.~Tran}, {\em A dictionary learning algorithm for
  compression and reconstruction of streaming data in preset order}, Discrete
  and Continuous Dynamical Systems-Series S (Print), 15 (2022).

\bibitem{baddoo2022kernel}
{\sc P.~J. Baddoo, B.~Herrmann, B.~J. McKeon, and S.~L. Brunton}, {\em Kernel
  learning for robust dynamic mode decomposition: linear and nonlinear
  disambiguation optimization}, Proceedings of the Royal Society A, 478 (2022),
  p.~20210830.

\bibitem{video_compression}
{\sc V.~Bhaskaran and K.~Konstantinides}, {\em Image and video compression
  standards: algorithms and architectures},  (1997).

\bibitem{brunton2016discovering}
{\sc S.~L. Brunton, J.~L. Proctor, and J.~N. Kutz}, {\em Discovering governing
  equations from data by sparse identification of nonlinear dynamical systems},
  Proceedings of the National Academy of Sciences, 113 (2016), pp.~3932--3937.

\bibitem{eigenvaluedecay}
{\sc C.-H. Chang and C.-W. Ha}, {\em On eigenvalues of differentiable positive
  definite kernels}, Integral Equations and Operator Theory, 33 (1999),
  pp.~1--7, \url{https://doi.org/10.1007/BF01203078},
  \url{https://doi.org/10.1007/BF01203078}.

\bibitem{streaming_compressed_sensing}
{\sc N.~M. Freris, O.~Öçal, and M.~Vetterli}, {\em Compressed sensing of
  streaming data}, in 2013 51st Annual Allerton Conference on Communication,
  Control, and Computing (Allerton), 2013, pp.~1242--1249,
  \url{https://doi.org/10.1109/Allerton.2013.6736668}.

\bibitem{Gurevich}
{\sc D.~R. Gurevich, P.~A.~K. Reinbold, and R.~O. Grigoriev}, {\em {Robust and
  optimal sparse regression for nonlinear PDE models}}, Chaos: An
  Interdisciplinary Journal of Nonlinear Science, 29 (2019), p.~103113,
  \url{https://doi.org/10.1063/1.5120861},
  \url{https://doi.org/10.1063/1.5120861},
  \url{https://arxiv.org/abs/https://pubs.aip.org/aip/cha/article-pdf/doi/10.1063/1.5120861/14622793/103113\_1\_online.pdf}.

\bibitem{hansen1999curve}
{\sc P.~C. Hansen}, {\em The l-curve and its use in the numerical treatment of
  inverse problems},  (1999).

\bibitem{streaming_DMD}
{\sc M.~S. Hemati, M.~O. Williams, and C.~W. Rowley}, {\em Dynamic mode
  decomposition for large and streaming datasets}, Physics of Fluids, 26
  (2014).

\bibitem{holmes2012turbulence}
{\sc P.~Holmes, J.~L. Lumley, G.~Berkooz, and C.~W. Rowley}, {\em Turbulence,
  coherent structures, dynamical systems and symmetry}, Cambridge university
  press, 2012.

\bibitem{PhysRevE.104.015206}
{\sc A.~A. Kaptanoglu, K.~D. Morgan, C.~J. Hansen, and S.~L. Brunton}, {\em
  Physics-constrained, low-dimensional models for magnetohydrodynamics:
  First-principles and data-driven approaches}, Phys. Rev. E, 104 (2021),
  p.~015206, \url{https://doi.org/10.1103/PhysRevE.104.015206},
  \url{https://link.aps.org/doi/10.1103/PhysRevE.104.015206}.

\bibitem{streaming_POD}
{\sc X.~Li, S.~Hulshoff, and S.~Hickel}, {\em An enhanced algorithm for online
  proper orthogonal decomposition and its parallelization for unsteady
  simulations}, Computers \& Mathematics with Applications, 126 (2022),
  pp.~43--59.

\bibitem{online_dictionary}
{\sc C.~Lu, J.~Shi, and J.~Jia}, {\em Online robust dictionary learning}, in
  2013 IEEE Conference on Computer Vision and Pattern Recognition, 2013,
  pp.~415--422, \url{https://doi.org/10.1109/CVPR.2013.60}.

\bibitem{messenger2021weakpde}
{\sc D.~A. Messenger and D.~M. Bortz}, {\em Weak sindy for partial differential
  equations}, Journal of Computational Physics, 443 (2021), p.~110525.

\bibitem{messenger2021weak}
{\sc D.~A. Messenger and D.~M. Bortz}, {\em Weak sindy: Galerkin-based
  data-driven model selection}, Multiscale Modeling \& Simulation, 19 (2021),
  pp.~1474--1497.

\bibitem{messenger2022online}
{\sc D.~A. Messenger, E.~Dall’Anese, and D.~Bortz}, {\em Online weak-form
  sparse identification of partial differential equations}, in Mathematical and
  Scientific Machine Learning, PMLR, 2022, pp.~241--256.

\bibitem{climate}
{\sc J.~Overpeck, G.~Meehl, S.~Bony, and D.~Easterling}, {\em Climate data
  challenges in the 21st century}, Science (New York, N.Y.), 331 (2011),
  pp.~700--2, \url{https://doi.org/10.1126/science.1197869}.

\bibitem{liftandlearn}
{\sc E.~Qian, I.-G. Farca\c{s}, and K.~Willcox}, {\em Reduced operator
  inference for nonlinear partial differential equations}, SIAM Journal on
  Scientific Computing, 44 (2022), pp.~A1934--A1959,
  \url{https://doi.org/10.1137/21M1393972},
  \url{https://doi.org/10.1137/21M1393972},
  \url{https://arxiv.org/abs/https://doi.org/10.1137/21M1393972}.

\bibitem{SCC.Rosenfeld.Kamalapurkar.ea2019a}
{\sc J.~A. Rosenfeld, R.~Kamalapurkar, B.~Russo, and T.~T. Johnson}, {\em
  Occupation kernels and densely defined {L}iouville operators for system
  identification}, in Proc. IEEE Conf. Decis. Control, Dec. 2019,
  pp.~6455--6460, \url{https://doi.org/10.1109/CDC40024.2019.9029337}.

\bibitem{rosenfeld2019occupation}
{\sc J.~A. Rosenfeld, B.~Russo, R.~Kamalapurkar, and T.~T. Johnson}, {\em The
  occupation kernel method for nonlinear system identification}, arXiv preprint
  arXiv:1909.11792,  (2019).

\bibitem{russo2022convergence}
{\sc B.~Russo and M.~P. Laiu}, {\em Convergence of weak-sindy surrogate
  models}, 2022, \url{https://arxiv.org/abs/2209.15573}.

\bibitem{Lyapunov}
{\sc T.~Savary}, {\em lyapynov 1.0.1},
  \url{https://pypi.org/project/lyapynov/}.

\bibitem{PhysRevE.96.023302}
{\sc H.~Schaeffer and S.~G. McCalla}, {\em Sparse model selection via integral
  terms}, Phys. Rev. E, 96 (2017), p.~023302,
  \url{https://doi.org/10.1103/PhysRevE.96.023302},
  \url{https://link.aps.org/doi/10.1103/PhysRevE.96.023302}.

\bibitem{streaming_SPOD}
{\sc O.~T. Schmidt and A.~Towne}, {\em An efficient streaming algorithm for
  spectral proper orthogonal decomposition}, Computer Physics Communications,
  237 (2019), pp.~98--109.

\bibitem{Lattice_boltzmann}
{\sc D.~V. Schroeder}, {\em {F}luid {D}ynamics {S}imulation}.
\newblock \url{https://physics.weber.edu/schroeder/fluids/}.

\bibitem{irregulart_integration}
{\sc N.~Shklov}, {\em Simpson's rule for unequally spaced ordinates}, The
  American Mathematical Monthly, 67 (1960), pp.~1022--1023,
  \url{http://www.jstor.org/stable/2309244} (accessed 2023-07-17).

\bibitem{tibshirani1996regression}
{\sc R.~Tibshirani}, {\em Regression shrinkage and selection via the lasso},
  Journal of the Royal Statistical Society: Series B (Methodological), 58
  (1996), pp.~267--288.

\bibitem{wu2021challenges}
{\sc Z.~Wu, S.~L. Brunton, and S.~Revzen}, {\em Challenges in dynamic mode
  decomposition}, Journal of the Royal Society Interface, 18 (2021),
  p.~20210686.

\bibitem{zhang2019online}
{\sc H.~Zhang, C.~W. Rowley, E.~A. Deem, and L.~N. Cattafesta}, {\em Online
  dynamic mode decomposition for time-varying systems}, SIAM Journal on Applied
  Dynamical Systems, 18 (2019), pp.~1586--1609.

\end{thebibliography}

\end{document}